\documentclass[10pt,journal,compsoc]{IEEEtran}
\ifCLASSOPTIONcompsoc
  \usepackage[nocompress]{cite}
\else
  \usepackage{cite}
\fi
\ifCLASSINFOpdf
\else
\fi
\usepackage{url}
\usepackage{amsmath}
\usepackage{amsthm}
\usepackage{amsfonts}
\usepackage{graphicx}
\usepackage{algorithm}
\usepackage{algorithmic}
\usepackage{bm}
\usepackage{color}
\usepackage{colortbl}
\usepackage{comment}
\usepackage{enumitem}
\usepackage{wrapfig}
\usepackage{booktabs}       
\newtheorem{definition}{Definition}
\newtheorem{theorem}{Theorem}
\newtheorem{corollary}{Corollary}
\newtheorem{remark}{Remark}
\usepackage{array}
\usepackage{multirow}
\usepackage[left]{lineno}
\usepackage{adjustbox}

\newcolumntype{P}[1]{>{\centering\arraybackslash}p{#1}}

\newcommand{\ab}[1]{\textbf{#1}}
\newcommand{\as}[1]{\underline{#1}}

\usepackage[normalem]{ulem}
\useunder{\uline}{\ul}{}

\begin{document}
\begin{sloppy}

%
\title{Permutation-equivariant and Proximity-aware Graph Neural Networks with Stochastic Message Passing}
%
%
%
%

\author{Ziwei~Zhang,
        Chenhao~Niu,
        Peng~Cui,
        Jian~Pei,~\IEEEmembership{Fellow,~IEEE,}
        Bo~Zhang,
        and~Wenwu~Zhu,~\IEEEmembership{Fellow,~IEEE}
\IEEEcompsocitemizethanks{
\IEEEcompsocthanksitem Z. Zhang, P. Cui, and W. Zhu are with the Department of Computer Science and Technology at Tsinghua University, Beijing 100084, China. Email: \{zwzhang,cuip,wwzhu\}@tsinghua.edu.cn
\IEEEcompsocthanksitem C. Niu is with the School of Computer Science at Carnegie Mellon University, Pittsburgh, PA 15213, USA. Work was done during his undergraduate study at Tsinghua Unversity. E-mail: cniu@andrew.cmu.edu
\IEEEcompsocthanksitem J. Pei is with the School of Computing Science at Simon Fraser University, Burnaby, BC V5A 1S6, Canada. E-mail: jpei@cs.sfu.ca
\IEEEcompsocthanksitem B. Zhang is with Tencent. E-mail:  nevinzhang@tencent.com.
\IEEEcompsocthanksitem Z. Zhang and C. Niu contributed equally to this work.
}
\thanks{Manuscript received June 01, 2021; revised January 14, 2022.}}

%
%

\markboth{Journal of \LaTeX\ Class Files,~Vol.~14, No.~8, August~2015}%
{Shell \MakeLowercase{\textit{et al.}}: Bare Demo of IEEEtran.cls for Computer Society Journals}
%



\IEEEtitleabstractindextext{%
\begin{abstract}
Graph neural networks (GNNs) are emerging machine learning models on graphs. Permutation-equivariance and proximity-awareness are two important properties highly desirable for GNNs. Both properties are needed to tackle some challenging graph problems, such as finding communities and leaders. In this paper, we first analytically show that the existing GNNs, mostly based on the message-passing mechanism, cannot simultaneously preserve the two properties. Then, we propose Stochastic Message Passing (SMP) model, a general and simple GNN to maintain both proximity-awareness and permutation-equivariance. In order to preserve node proximities, we augment the existing GNNs with stochastic node representations. We theoretically prove that the mechanism can enable GNNs to preserve node proximities, and at the same time, maintain permutation-equivariance with certain parametrization. We report extensive experimental results on ten datasets and demonstrate the effectiveness and efficiency of SMP for various typical graph mining tasks, including graph reconstruction, node classification, and link prediction.
\end{abstract}

}

\maketitle

\IEEEdisplaynontitleabstractindextext

%
\IEEEpeerreviewmaketitle


%
%
%
%

 

\IEEEraisesectionheading{\section{Introduction}\label{sec:introduction}}
\IEEEPARstart{G}{raph} neural networks (GNNs), as generalizations of neural networks for learning on graph data, have enjoyed successes in many applications, such as social recommendation~\cite{NIPS2019_8808}, physical simulation~\cite{kipf2018neural}, and protein interaction prediction~\cite{hamilton2017inductive}.  The existing GNNs are mostly based on the message-passing mechanism~\cite{gilmer2017neural}.

A fundamental property well preserved by the message-passing GNNs is permutation-equivariance, i.e., if we randomly permutate the IDs of nodes while maintaining the graph structure unchanged, the representations of nodes in those GNNs are permutated accordingly. Mathematically, permutation-equivariance reflects one basic symmetric group of graph structures.  Permutation-equivariance is highly useful for many graph mining tasks, such as node or graph classification~\cite{keriven2019universal,maron2018invariant}. As another important property, pairwise proximities between nodes are crucial for some other graph mining tasks, such as link prediction and community detection~\cite{hu2020open,you2019position}. Some GNNs, such as Position-aware GNN (P-GNN)~\cite{you2019position}, have specifically designed mechanisms to ensure proximity-awareness. 

In many applications of GNNs, both proximity-awareness and permutation-equivariance are indispensable. Consider mining communities and leaders in graphs. Figure~\ref{fig:example} shows a toy example for illustration. Figure~\ref{fig:example}(a) shows the communities and Figure~\ref{fig:example}(b) shows the nodes categorized by the k-core number centrality. To discover the communities, proximity-awareness is essential since the nodes in the same community are tightly connected and have large proximities.  Permutation-equivariance helps to measure the centrality because most centrality measurements are permutation-equivariant by definition~\cite{borgatti2005centrality}.

\begin{figure}[t] 
	\centering
	\includegraphics[width=3.1in]{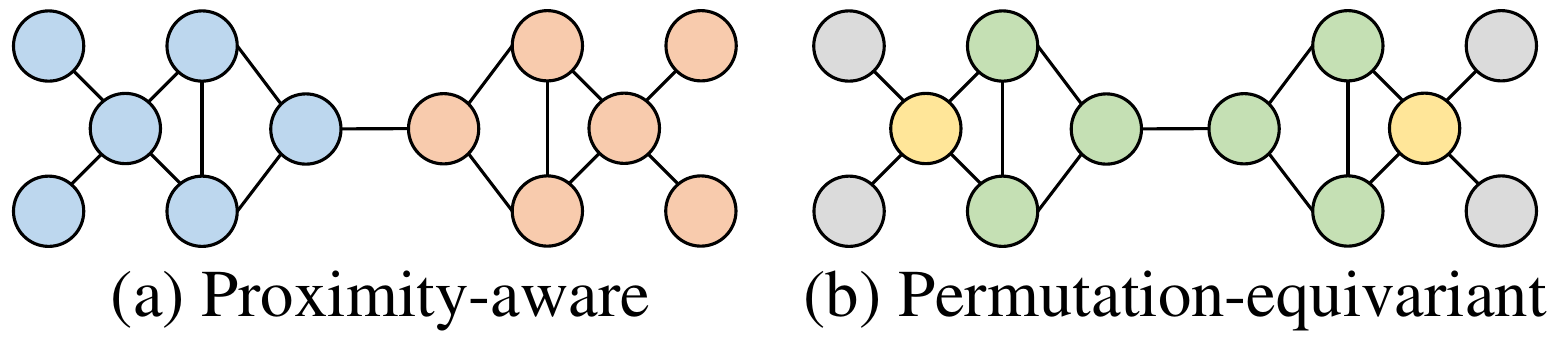} 
	\caption{A toy example of illustrating finding communities and leaders. Labels are shown in different colors. (a) The two communities discovered by spectral clustering, in which proximity-awareness is essential. (b) The node labels correspond to the k-core number, a type of node centrality. Permutation-equivariance is important for the task.}
	\label{fig:example} 
\end{figure}

Do the existing GNNs, which are built on message-passing, honor both proximity-awareness and permutation-equivariance?  Surprisingly and unfortunately, the answer is no.  
We show that proximity-awareness and permutation-equivariance are incompatible in the exiting GNNs (see Theorem~\ref{thm:walksim}). This deficiency in the existing GNNs is particularly irritating since, for the same task, different datasets may rely on the two properties to different extents.
Taking link prediction as an example, we observe that permutation-equivariant GNNs such as GCN~\cite{kipf2017semi} or GAT~\cite{velickovic2018graph} show better performance than P-GNN in coauthor graphs, but perform worse in biological graphs (see Section~\ref{sec:exp-linkprediction} for details). A work in drug repurposing for Covid-19~\cite{gysi2020network} shows a similar dilemma: proximity-aware methods and permutation-equivariant GNNs discover completely different drug candidates. 

Can we develop a general GNN that is proximity-aware and also maintains permutation-equivariance?  
In this paper, we propose Stochastic Message Passing (SMP)\footnote{Code is available at \url{https://github.com/NiuChH/SMP}.}, a general and simple GNN to preserve both proximity-awareness and permutation-equivariance properties. In order to preserve node proximities, we augment the existing GNNs with stochastic node representations. We theoretically prove that the mechanism can enable GNNs to preserve node proximities (see Theorems~\ref{thm:walksim2} and~\ref{thm:walksim3}).
At the same time, SMP is equivalent to a permutation-equivariant GNN with certain parametrization and thus is at least as powerful as those GNNs in permutation-equivariant tasks (see Remark~\ref{thm:permu}). Therefore, SMP is general and flexible in handling both proximity-aware and permutation-equivariant tasks, which is also demonstrated by our extensive experimental results.
Besides, owing to the simple structure, SMP is computationally efficient, with a running time roughly the same as the simplest GNNs, such as SGC~\cite{wu2019simplifying}, and is at least an order of magnitude faster than P-GNN on large graphs.
Our ablation studies further show that a linear instantiation of SMP is expressive enough as adding extra non-linearities does not lift the performance of SMP on the majority of datasets. 
Our contributions are summarized as follows.

\begin{itemize}
	\item We propose Stochastic Message Passing (SMP), a simple and general GNN to handle both proximity-aware and permutation-equivariant graph tasks.
	\item We prove that SMP has theoretical guarantees in preserving walk-based proximities and is as powerful as the existing GNNs in permutation-equivariant tasks.
	\item Extensive experimental results demonstrate the effectiveness and efficiency of SMP. We show that a linear SMP instantiation is expressive enough on the majority of datasets.
\end{itemize}

The rest of the paper is organized as follows.  We review related work in Section~\ref{sec:related-work}. In Section~\ref{sec:pre}, we show the incompatibility between walk-based proximity and permutation-equivariance in message-passing GNNS.  We develop our proposed Stochastic Message Passing in Section~\ref{sec:SMPGNN}.  We report an extensive experimental study in Section~\ref{sec:exp}, and conclude the paper in Section~\ref{sec:con}. We provide additional experiments, details for reproducibility, and proofs in the appendix.

\section{Related Work}\label{sec:related-work}
We briefly review GNNs, the permutation-equivariance property, and the proximity-awareness property. We refer readers to \cite{zhang2020deep} for a comprehensive survey. 

The earliest GNNs adopt a recursive definition of node states~\cite{scarselli2008graph,gori2005new} or a contextual realization~\cite{micheli2009neural}. GGS-NNs~\cite{li2016gated} replace the recursive definition with recurrent neural networks (RNNs). Spectral GCNs~\cite{bruna2014spectral} define graph convolutions using graph signal processing~\cite{shuman2013emerging} with ChebNet~\cite{defferrard2016convolutional} and GCN~\cite{kipf2017semi} approximating the spectral filters using a low-order Chebyshev polynomial and the first-order polynomial, respectively. MPNNs~\cite{gilmer2017neural}, GraphSAGE~\cite{hamilton2017inductive}, and MoNet~\cite{monti2017geometric} are proposed as general frameworks by characterizing GNNs with a message-passing function and an updating function. More advanced variants such as GAT~\cite{velickovic2018graph}, JK-Nets~\cite{xu2018representation}, GIN~\cite{xu2019powerful}, and GraphNets~\cite{battaglia2018relational} follow these frameworks.

Li~\textit{et~al.}~\cite{li2018deeper}, Xu~\textit{et~al.}~\cite{xu2019powerful}, Morris~\textit{et al.}~\cite{morris2019weisfeiler}, and Maron~\textit{et al.}~\cite{maron2019provably} show the connection between GNNs and the Weisfeiler-Lehman algorithm~\cite{shervashidze2011weisfeiler} of graph isomorphism tests, in which permutation-equivariance holds a key constraint. Further, Maron~\textit{et~al.}~\cite{maron2018invariant} and Keriven~\textit{et~al.}~\cite{keriven2019universal} analyze the permutation-equivariance property of GNNs theoretically. To date, most of the existing GNNs are permutation-equivariant and are not proximity-aware. One exception is P-GNN~\cite{you2019position}, which proposes to capture the positions of nodes using the relative distance between the target node and some randomly chosen anchor nodes. However, P-GNN cannot satisfy permutation-equivariance and is computationally expensive. Concurrent to our work, some studies propose to use position encodings to enhance GNNs in preserving graph structures~\cite{cui2021positional,dwivedi2020benchmarking,beani2021directional,zhang2021eigen}. Most of these methods rely on eigenvectors of a graph matrix, which are computationally expensive. Our proposed stochastic node representations can also be regarded as a type of position encoding while being extremely simple yet efficient. 

In order to enhance the expressive power of GNNs in graph isomorphism tests and also motivated by the literature on distributed computing~\cite{angluin1980local}, some studies suggest assigning unique node identifiers for GNNs~\cite{loukas2019graph}, such as one-hot IDs~\cite{murphy2019relational} or random numbers~\cite{dasoulas2019coloring,sato2020random,corso2020principal}.
For example, Sato~\textit{et~al.}~\cite{sato2020random} show that random numbers can enhance GNNs in tackling two graph-based NP problems with a theoretical guarantee, namely the minimum dominating set and the maximum matching problem. Fey~\textit{et~al.}~\cite{fey2019deep} empirically show the effectiveness of random features in the graph matching problem.
Concurrent to our work, RNI~\cite{RNI} shows that GNNs with random node features are universal in theory.
Our work here, which also adopts stochastic node representations, differs in that we systematically study how to preserve permutation-equivariance and proximity-awareness simultaneously in a simple yet effective framework, a novel topic distinct from those existing studies. Besides, we theoretically prove that our proposed method can preserve walk-based proximities. We also demonstrate the effectiveness of our method on large-scale benchmarks for both node- and edge-level tasks, while no similar results are reported in the literature.

Another line of research is tackling the over-smoothing problem~\cite{li2018deeper,oono2020graph} and developing deep GNNs~\cite{li2019deepgcns}. Since these studies are orthogonal to our paper, we expect these strategies to also work for our proposed SMP.

The design of our method is also inspired by the literature on random projection for dimensionality reduction~\cite{vempala2005random}.  To the best of our knowledge, we are the first to study random projection in the scope of GNNs. More remotely, our definition of node proximities is inspired and inherited from graph kernels~\cite{gartner2003graph,borgwardt2005shortest}, network embedding~\cite{perozzi2014deepwalk,grover2016node2vec}, and the general studies of graphs~\cite{newman2018networks}.

\section{Message-passing GNNs and Analyses}\label{sec:pre}
In this section, we first introduce preliminaries of message-passing GNNs, walk-based proximities, and permutation-equivariance. Then, we show the incompatibility between  proximity-awareness and permutation-equivariance in the existing GNNs. 

We consider a graph $G=\left(\mathcal{V},\mathcal{E},\mathbf{F}\right)$ where $\mathcal{V} = \left\{v_1,...,v_N\right\}$ is a set of  $N = \left| \mathcal{V}\right|$ nodes, $\mathcal{E} \subseteq \mathcal{V} \times \mathcal{V}$ is a set of $M = \left| \mathcal{E}\right|$ edges, and $\mathbf{F} \in \mathbb{R}^{N \times d_0}$ is a matrix of $d_0$ node features. Denote by $\mathbf{A}$ the adjacency matrix, and by $\mathbf{A}_{i,:}$, $\mathbf{A}_{:,j}$, and $\mathbf{A}_{i,j}$, respectively, the $i^{th}$ row, the $j^{th}$ column and an element in the matrix. In this paper, we assume unweighted and undirected graphs. The neighborhood of node $v_i$ is denoted by $\mathcal{N}_i$. Let $\tilde{\mathcal{N}}_i = \mathcal{N}_i \cup \left\{v_i\right\}$. 

The existing GNNs usually follow a message-passing framework \cite{gilmer2017neural}, where a neighborhood aggregation function $\text{AGG}(\cdot)$ and an updating function $\text{UPDATE}(\cdot)$ are adopted in the $l^{th}$ layer:
\begin{equation}
	\begin{gathered}\label{eq:mpnn}
		\mathbf{m}_i^{(l)} = \text{AGG}\left(
		\left\{ \mathbf{h}_j^{(l)}, \mathbf{h}_i^{(l)}, \mathbf{e}_{i,j}, \forall j \in \tilde{\mathcal{N}}_i \right\}, \right) \\
		\mathbf{h}_i^{(l+1)} = \text{UPDATE}\left( \left[\mathbf{h}_i^{(l)},  \mathbf{m}_i^{(l)}\right]\right),
	\end{gathered}
\end{equation}
where $\mathbf{h}_i^{(l)} \in \mathbb{R}^{d_l}$ is the representation of node $v_i$ at the $l^{th}$ layer, $d_l$ is the dimensionality, $\mathbf{e}_{i,j}$ is the edge feature when available, and $\mathbf{m}_i^{(l)}$ are the messages. We also denote by $\mathbf{H}^{(l)} = [\mathbf{h}^{(l)}_1,...,\mathbf{h}^{(l)}_N]$ and $\left[\cdot,\cdot\right]$ the concatenation operation. The node representations are initialized as node features, i.e., $\mathbf{H}^{(0)} = \mathbf{F}$. We represent a GNN following Eq.~\eqref{eq:mpnn} with $L$ layers by a parameterized function as follows\footnote{Since the final layer of GNNs is task-specific, e.g., a softmax layer for node classification or a readout layer for graph classification, we only consider the GNN architecture to its last hidden layer.}:
\begin{equation}\label{eq:GNN}
	\mathbf{H}^{(L)} = \mathcal{F}_{\text{GNN}}(\mathbf{A},\mathbf{F}; \mathbf{W}),
\end{equation}
where $\mathbf{H}^{(L)}$ is node representation learned by the GNN and $\mathbf{W}$ represents all the parameters.

A key property of GNNs is permutation-equivariance.
\begin{definition}[Permutation-equivariance]\label{def:equi}
	Consider a graph $G=\left(\mathcal{V},\mathcal{E},\mathbf{F}\right)$ and any permutation $\mathcal{P}:\mathcal{V} \rightarrow \mathcal{V}$ so that $G^\prime=\left(\mathcal{V},\mathcal{E}^\prime,\mathbf{F}^\prime\right)$ has an adjacency matrix $\mathbf{A}^\prime = \mathbf{P} \mathbf{A} \mathbf{P}^T$ and a feature matrix $\mathbf{F}^\prime = \mathbf{P} \mathbf{F}$, where $\mathbf{P} \in \left\{0,1\right\}^{N \times N}$ is the permutation matrix, i.e., $\mathbf{P}_{i,j}$ = 1 iff $\mathcal{P}(v_i) = v_j$. A GNN satisfies permutation-equivariance if the node representations for $G$ and $G^\prime$ are equivariant with respect to $\mathbf{P}$, i.e.,
	\begin{equation}
		\mathbf{P} \mathcal{F}_{\text{GNN}}(\mathbf{A},\mathbf{F}; \mathbf{W}) =
		\mathcal{F}_{\text{GNN}}(\mathbf{P} \mathbf{A} \mathbf{P}^T, \mathbf{P} \mathbf{F}; \mathbf{W}).
	\end{equation}
\end{definition}
It is well-known that GNNs following Eq.~\eqref{eq:mpnn} satisfy permutation-equivariance~\cite{maron2018invariant}.

\begin{definition}[Automorphism]\label{def:auto}
A graph $G$ is said to have (non-trivial) automorphism if there exists a non-identity permutation matrix $\mathbf{P} \neq \mathbf{I}_N$ so that $\mathbf{A} = \mathbf{P} \mathbf{A} \mathbf{P}^T$ and $\mathbf{F} = \mathbf{P} \mathbf{F}$, i.e., the graph has a non-trivial isomorphism to itself. We denote by $\mathcal{C}_G= \bigcup_{\mathbf{P}\neq \mathbf{I}_N} \left\{(i,j)| \mathbf{P}_{i,j} \neq 0, i \neq j\right\}$ the corresponding automorphic node pairs.
\end{definition}

Using Definition~\ref{def:equi} and \ref{def:auto}, we immediately have the following corollary.
\begin{corollary}\label{cor:automorphic}
If a graph has a non-trivial automorphism, a permutation-equivariant GNN produces identical node representations for automorphic node pairs
\begin{equation}
    \mathbf{h}^{(L)}_{i} = \mathbf{h}^{(L)}_{j}, \forall (i,j) \in \mathcal{C}_G.
\end{equation}
\end{corollary}

Since node representations are used for downstream tasks, Corollary~\ref{cor:automorphic} shows that permutation-equivariant GNNs cannot differentiate automorphic node pairs. 
A direct consequence is that permutation-equivariant GNNs cannot preserve walk-based proximities between pairs of nodes.

\begin{definition}[Walk-based Proximities]\label{def:proximity}
	For a given graph $G=\left(\mathcal{V},\mathcal{E},\mathbf{F}\right)$, denote by matrix $\mathbf{S}\in\mathbb{R}^{N\times N}$ the walk-based proximities between pairs of nodes, defined by
	\begin{equation}
		\mathbf{S}_{i,j} = \mathcal{S}\left (\left\{v_i \leadsto v_j \right\} \right),
	\end{equation}
	where $\left\{v_i \leadsto v_j\right\}$ represents the set of walks from node $v_i$ to $v_j$ and $\mathcal{S}(\cdot)$ is a real-valued function. The length of a walk-based proximity is the maximum length of all the walks of the proximity. 
\end{definition}
Typical examples of walk-based proximities include the shortest distance~\cite{you2019position}, the high-order proximities (a sum of walks weighted by their lengths)~\cite{zhang2018arbitrary}, and random walk probabilities~\cite{klicpera2019predict}.

\begin{definition}
For a given walk-based proximity, a GNN is said to preserve the proximity if there exists a decoder function $\mathcal{F}_{\text{de}}(\cdot)$ such that for any graph $G=\left(\mathcal{V},\mathcal{E},\mathbf{F}\right)$, there exist parameters $\mathbf{W}_G$ so that $\forall \epsilon > 0$:
\begin{equation}
     \left| \mathbf{S}_{i,j} - \mathcal{F}_{\text{de}}\left(\mathbf{H}^{(L)}_{i,:},\mathbf{H}^{(L)}_{j,:};\mathcal{S}(\cdot)\right) \right| < \epsilon,
\end{equation}
where
\begin{equation}
    \mathbf{H}^{(L)} = \mathcal{F}_{\text{GNN}}(\mathbf{A},\mathbf{F}; \mathbf{W}_G).
\end{equation}
For notation convenience, we also write the decoder function as $\mathcal{F}_{\text{de}}\left(\mathbf{H}^{(L)}_{i,:},\mathbf{H}^{(L)}_{j,:} \right)$ when there is no ambiguity regarding $\mathcal{F}(\cdot)$.
\end{definition}

The definition applies to any GNN architecture
as long as it fits Eq.~\eqref{eq:mpnn}.  Moreover, in the definition, we only constrain the inputs of the decoder function to be node representations $\mathbf{H}$ and the proximity function $\mathcal{S}(\cdot)$, but we do not constrain the form of the decoder function. In other words, the decoder function can be arbitrarily sophisticated, e.g., deep neural networks with a sufficient number of layers and hidden units. Now we are ready to present the incompatibility.

\begin{theorem}\label{thm:walksim}
	For any walk-based proximity function $\mathcal{S}(\cdot)$ satisfying Definition~\ref{def:proximity}, a permutation-equivariant GNN cannot preserve $\mathcal{S}(\cdot)$, except for the trivial situation where all node pairs have the same proximity, i.e., $\forall i,j$, $\mathbf{S}_{i,j} = c$, and $c$ is a constant.\footnote{Proposition~1 in P-GNN~\cite{you2019position} can be regarded as a special case of Theorem~\ref{thm:walksim} using the shortest distance proximity.}
\end{theorem}
\begin{proof}
	We prove by contradiction. Assume there exists a non-trivial $\mathcal{S}(\cdot)$ that can be preserved by a permutation-equivariant GNN. Consider any graph $G=\left(\mathcal{V},\mathcal{E},\mathbf{F}\right)$. We will construct a graph $G^\prime$ with automorphism from $G$ so that any GNN cannot preserve $\mathcal{S}(\cdot)$ on $G^\prime$. Specifically, let $N = \left| \mathcal{V} \right|$. We create $G^\prime=\left(\mathcal{V}^\prime,\mathcal{E}^\prime,\mathbf{F}^\prime\right), \left| \mathcal{V}^\prime \right| = 2 N$, such that
	\begin{displaymath}\label{eq:construct1}
		\begin{gathered}
			\mathcal{E}^\prime_{i,j} = \left\{
			\begin{aligned}
				& \mathcal{E}_{i,j} &\text{ if } i\leq N, j\leq N \\
				& \mathcal{E}_{i-N,j-N} &\text{ if } i > N, j > N \\
				& 0 &\text{ else } \\
			\end{aligned} \right. \\
			\mathbf{F}^\prime_{i,:} = \left\{
			\begin{aligned}
				& \mathbf{F}_{i,:} &\text{ if } i\leq N \\
				& \mathbf{F}_{i - N,:} &\text{ if } i > N \\
			\end{aligned} \right. .
		\end{gathered}
	\end{displaymath}

Basically, we generate two ``copies'' of the original graph, one indexing from $1$ to $N$, and the other one from $N+1$ to $2N$. By assumption, there exists a permutation-equivariant GNN that can preserve $\mathcal{S}(\cdot)$ in $G^\prime$. Let the node representations for such a GNN as $\mathbf{H}^{\prime(L)} = \mathcal{F}_{\text{GNN}}(\mathbf{A}^\prime,\mathbf{F}^\prime; \mathbf{W}_{G^\prime})$. It is easy to see that node $v_i^\prime$ and $v_{i+N}^\prime$ in $G^\prime$ form an automorphic node pair. According to Corollary~\ref{cor:automorphic}, their representations are identical, i.e.,
\begin{equation}
	\mathbf{H}^{\prime(L)}_{i,:} = \mathbf{H}^{\prime(L)}_{i + N,:}, \forall i \leq N.
\end{equation}

Note that there exists no walk from the two copies, i.e. $\left\{v_i^\prime \leadsto v_j^\prime\right\} = \left\{v_j^\prime \leadsto v_i^\prime\right\} = \emptyset, \forall i \leq N, j >N$.
As a result, for  $\forall i \leq N, j\leq N, \forall \epsilon > 0$, we have:
\begin{scriptsize}
	\begin{displaymath}\label{eq:disinequality}
		\begin{split}
			\left| \mathbf{S}_{i,j} - \mathcal{S}(\emptyset) \right|  \leq \left| \mathbf{S}_{i,j} -  \mathcal{F}_{\text{de}}\left(\mathbf{H}^{\prime(L)}_{i,:},\mathbf{H}^{\prime(L)}_{j,:}\right) \right| + \left| \mathcal{S}(\emptyset) - \mathcal{F}_{\text{de}}\left(\mathbf{H}^{\prime(L)}_{i,:},\mathbf{H}^{\prime(L)}_{j,:}\right) \right| \\
			= \left| \mathbf{S}_{i,j} -  \mathcal{F}_{\text{de}}\left(\mathbf{H}^{\prime(L)}_{i,:},\mathbf{H}^{\prime(L)}_{j,:}\right) \right| + \left| \mathbf{S}_{i,j+N} - \mathcal{F}_{\text{de}}\left(\mathbf{H}^{\prime(L)}_{i,:},\mathbf{H}^{\prime(L)}_{j+N,:}\right) \right| < 2\epsilon.
		\end{split}
	\end{displaymath}
\end{scriptsize}

We can prove the same for $\forall i > N, j > N$. The equation naturally holds if $i \leq N, j> N$ or $i > N, j \leq N$, since $\left\{v_i^\prime \leadsto v_j^\prime\right\} = \emptyset$. Combining the results, we have $\forall \epsilon > 0, \forall i, j$,  $\left| \mathbf{S}_{i,j} - \mathcal{S}(\emptyset) \right| < 2 \epsilon$. Since $\epsilon$ can be arbitrarily small, the equation shows that all node pairs have the same proximity $c = \mathcal{S}(\emptyset)$. In other words, $\mathcal{S}(\cdot)$ is a trivial situation.  A contradiction.
\end{proof}

An alternative proof by constructing connected graphs as contradictions is provided in Appendix~\ref{sec:proof:alter}.

Since walk-based proximities are rather general and widely used in many graph mining tasks such as link prediction, Theorem~\ref{thm:walksim} shows that the existing permutation-equivariant GNNs cannot handle these tasks well.

\section{Stochastic Message Passing}\label{sec:SMPGNN}
In this section, we develop our stochastic message passing model.  We first describe our framework, and then explore a linear implementation and non-linear extensions.
\subsection{Stochastic Message Passing Framework}\label{sec:SMP}
Theorem~\ref{thm:walksim} indicates that a major shortcoming of permutation-equivariant GNNs is that they cannot differentiate automorphic node pairs. To solve that problem, we need to introduce some mechanism as ``symmetry breaking'', i.e., to enable GNNs to distinguish symmetric nodes. To achieve this goal, we sample a stochastic matrix $\mathbf{E} \in \mathbb{R}^{N \times d}$, where each element follows an i.i.d. normal distribution $\mathcal{N}(0,1)$. We leave exploring other possible stochastic signals besides Gaussian distributions as future works. The stochastic matrix can provide signals to distinguish the nodes because they are randomly sampled without being affected by the graph automorphism. In fact, we can easily calculate that the Euclidean distance between two stochastic signals divided by a constant $\sqrt{2}$ follows a chi distribution $\chi_d$, that is,
\begin{equation}
	\frac{1}{\sqrt{2}}\left| \mathbf{E}_{i,:}  - \mathbf{E}_{j,:}\right| \sim \chi_d, \forall i,j.
\end{equation}

When $d$ is reasonably large, e.g., $d > 20$, the probability of two signals being close is very low.
Then, inspired by the message-passing framework, we apply a GNN on the stochastic matrix so that the nodes can exchange information of the stochastic signals,
\begin{equation}\label{eq:stochasticrep}
	\tilde{\mathbf{E}} = \mathcal{F}_{\text{GNN}}\left(\mathbf{A},\mathbf{E}; \mathbf{W} \right).
\end{equation}

We call $\tilde{\mathbf{E}}$ the stochastic representation of nodes. By the message-passing on the stochastic signals, $\tilde{\mathbf{E}}$ can be used to preserve node proximities (will be shown in Theorem~\ref{thm:walksim2} and Theorem~\ref{thm:walksim3} in a moment).
To still allow our model to utilize node features, we concatenate $\tilde{\mathbf{E}}$ with node representations from another GNN with node features as inputs. That is,
\begin{equation}\label{eq:SMPframe}
	\begin{gathered}
		\mathbf{H} = \mathcal{F}_{\text{output}} ( [\tilde{\mathbf{E}}, \mathbf{H}^{(L)} ]) \\
		\tilde{\mathbf{E}} = \mathcal{F}_{\text{GNN}}\left(\mathbf{A},\mathbf{E}; \mathbf{W} \right),\mathbf{H}^{(L)} = \mathcal{F}_{\text{GNN}^\prime}(\mathbf{A},\mathbf{F}; \mathbf{W}^\prime) ,
	\end{gathered}
\end{equation}
where $\mathcal{F}_{\text{output}}(\cdot)$ is an aggregation function, such as a linear function or simply the identity mapping. In a nutshell, our proposed method augments the existing GNNs with a stochastic representation learned by message-passings to differentiate different nodes and preserve node proximities.

There is also a delicate choice worthy mentioning, i.e., whether the stochastic matrix $\mathbf{E}$ is fixed or resampled in each epoch. On the one hand, by fixing $\mathbf{E}$, the model can learn to memorize the stochastic representation and distinguish different nodes, but with the cost of being unable to handle nodes not seen during training. On the other hand, by resampling $\mathbf{E}$ in each epoch, the model can have a better generalization ability since the model cannot simply remember one specific stochastic matrix. However, the node representations are not fixed (but pairwise proximities are preserved; see Theorem~\ref{thm:walksim2}). In these cases, $\tilde{\mathbf{E}}$ is more capable of handling pairwise tasks such as link prediction or pairwise node classification.

In this paper, we fix $\mathbf{E}$ for transductive datasets and resample $\mathbf{E}$ for inductive datasets (see Section~\ref{sec:exp-setup} for the experimental settings and Section~\ref{sec:exp-ablation} for an ablation study of 
this design).

\textbf{Time Complexity} From Eq.\eqref{eq:SMPframe}, the time complexity of our framework mainly depends on the two GNNs in learning the stochastic and permutation-equivariant node representations. In this paper, we instantiate these two GNNs using simple message-passing GNNs, such as GCN~\cite{kipf2017semi} and SGC~\cite{wu2019simplifying} (see Section~\ref{sec:linearins} and Section~\ref{sec:nonlinearins}). Thus, the time complexity of our method is the same as those models employed, which is $O(M)$, i.e., linear with respect to the number of edges. We also empirically compare the running time of different models in Section~\ref{sec:runtime}. Besides, GNN acceleration schemes such as sampling~\cite{chen2018stochastic,chen2018fastgcn,huang2018adaptive} or partitioning the graph~\cite{chiang2019cluster} can be directly applied to our framework.

\subsection{A Linear Instantiation}\label{sec:linearins}
Based on the general framework in Eq.~\eqref{eq:SMPframe}, let us explore its minimum model instantiation, i.e., a linear model. 

Specifically, inspired by Simplified Graph Convolution (SGC) \cite{wu2019simplifying}, we adopt a linear message-passing for both GNNs, i.e.,
\begin{equation}\label{eq:SMP}
	\mathbf{H} = \mathcal{F}_{\text{output}} ( [\tilde{\mathbf{E}}, \mathbf{H}^{(L)} ]) = \mathcal{F}_{\text{output}} ([\tilde{\mathbf{A}}^{K}\mathbf{E},\tilde{\mathbf{A}}^{K}\mathbf{F}] ),
\end{equation}
where $\tilde{\mathbf{A}} = (\mathbf{D} + \mathbf{I})^{-\frac{1}{2}} (\mathbf{A} + \mathbf{I}) (\mathbf{D} + \mathbf{I})^{-\frac{1}{2}}$ is the normalized graph adjacency matrix with self-loops proposed in GCN~\cite{kipf2017semi}, $\mathbf{I}$ is the identity matrix, and $K$ is the number of propagation steps. We also set $\mathcal{F}_{\text{output}}(\cdot)$ in Eq.~\eqref{eq:SMP} to a linear mapping or identity mapping. 

Elegantly, this simple SMP instantiation has a theoretical guarantee on preserving walk-based proximities.
\begin{theorem}\label{thm:walksim2}
	An SMP in Eq.~\eqref{eq:SMP} can preserve the walk-based proximity $\tilde{\mathbf{A}} ^{K}(\tilde{\mathbf{A}} ^{K})^T$ with high probability if the dimensionality of the stochastic matrix $d$ is sufficiently large,
	i.e., $\forall \epsilon > 0$ and $\delta > 0$, $\exists$ $d_0$ so that for any $d > d_0$,
	\begin{equation}
		P \left( \left| \mathbf{S}_{i,j} - \mathcal{F}_{\text{de}}\left(\mathbf{H}_{i,:},\mathbf{H}_{j,:}\right) \right| < \epsilon \right) > 1 - \delta,
	\end{equation}
	where $\mathbf{H}$ is the node representation obtained from SMP in Eq.~\eqref{eq:SMP}. The result holds for any stochastic matrix, no matter whether $\mathbf{E}$ is fixed or resampled during each epoch.
\end{theorem}

\begin{proof}
	Our proof is mostly based on the random projection theory. First, since we show in Theorem~\ref{thm:walksim} that the permutation-equivariant representations cannot preserve any walk-based proximity, here we develop our proof assuming $\mathbf{H} = \tilde{\mathbf{E}}$. This can be easily achieved in the model by ignoring $\mathbf{H}^{(L)}$ in $\mathcal{F}_{\text{output}} ( [\tilde{\mathbf{E}},\mathbf{H}^{(L)}])$. For example, if we set $\mathcal{F}_{\text{output}}(\cdot)$ as a linear function, the model can learn to set the corresponding weights for $\mathbf{H}^{(L)}$ as all-zeros and weights for $\tilde{\mathbf{E}}$ as an identity matrix.
	
	We set the decoder function as a normalized inner product
	\begin{equation}
		\mathcal{F}_{\text{de}}\left(\mathbf{H}_{i,:},\mathbf{H}_{j,:}\right) = \frac{1}{d} \mathbf{H}_{i,:} \mathbf{H}_{j,:}^T = \frac{1}{d} \tilde{\mathbf{E}}_{i,:} \tilde{\mathbf{E}}_{j,:}^T.
	\end{equation}
	Let $\mathbf{a}_i = \tilde{\mathbf{A}}^{K}_{i,:}$. Recall $\tilde{\mathbf{E}} = \tilde{\mathbf{A}}^{K}\mathbf{E}$. Then, we have
	\begin{small}
	\begin{displaymath}
		\left| \mathbf{S}_{i,j} - \mathcal{F}_{\text{de}}\left(\mathbf{H}_{i,:},\mathbf{H}_{j,:}\right) \right| = | \mathbf{a}_i \mathbf{a}_j^T -  \frac{1}{d} \tilde{\mathbf{E}}_{i,:} \tilde{\mathbf{E}}_{j,:}^T| = |\mathbf{a}_i \mathbf{a}_j^T - \frac{1}{d} \mathbf{a}_i \mathbf{E}\mathbf{E}^T\mathbf{a}_j^T |.
	\end{displaymath}
	\end{small}
	Since $\mathbf{E}$ is a Gaussian random matrix, using the Johnson-Lindenstrauss lemma \cite{vempala2005random} (in the inner product preservation form, e.g., see Corollary 2.1 and its proof in \cite{CSMC35900}), $\forall 0 < \epsilon^\prime < \frac{1}{2}$, we have
	\begin{small}
	\begin{displaymath}
		P\left( |\mathbf{a}_i \mathbf{a}_j^T - \frac{1}{d}\mathbf{a}_i \mathbf{E}\mathbf{E}^T \mathbf{a}_j^T | \leq \frac{\epsilon^\prime}{2} (\left\| \mathbf{a}_i \right\| + \left\| \mathbf{a}_j\right\|) \right) >  1-  4 e^{-\frac{({\epsilon^{\prime}}^2 - {\epsilon^{\prime}}^3)d}{4}}.
	\end{displaymath}
	\end{small}
	
	By setting $\epsilon^\prime = \frac{\epsilon}{\max_i \left\| \mathbf{a}_i \right\|}$, we have $\epsilon > \frac{\epsilon^\prime}{2} (\left\| \mathbf{a}_i \right\| + \left\| \mathbf{a}_j\right\|)$ and
	\begin{small}
	\begin{displaymath}
		P\left( \left| \mathbf{S}_{i,j} - \mathcal{F}_{\text{de}}\left(\mathbf{H}_{i,:},\mathbf{H}_{j,:}\right) \right|  < \epsilon\right) >  1-  4 e^{-\frac{({\frac{\epsilon}{\max_i \left\| \mathbf{a}_i \right\|}}^2 - {\frac{\epsilon}{\max_i \left\| \mathbf{a}_i \right\|}}^3)d}{4}},
	\end{displaymath}
	\end{small}
	which leads to the theorem by solving and setting $d_0$ as follows.
	\begin{displaymath}
		4 e^{-\frac{({\frac{\epsilon}{\max_i \left\| \mathbf{a}_i \right\|}}^2 - {\frac{\epsilon}{\max_i \left\| \mathbf{a}_i \right\|}}^3)d_0}{4}} = \delta \Rightarrow
		d_0 = \frac{4 \log\frac{4}{\delta} \left( \max_i \left\| \mathbf{a}_i \right\|\right)^3}{ \epsilon^2 \max_i \left\| \mathbf{a}_i \right\| - \epsilon^3}.
	\end{displaymath}
\end{proof}

Next, we show that SMP is equivalent to a permutation-equivariant GNN with certain parametrization.
\begin{remark}\label{thm:permu}
	Suppose we adopt $\mathcal{F}_{\text{output}}(\cdot)$ as a linear function with the output dimensionality same as $\mathcal{F}_{\text{GNN}^\prime}$. Then, Eq.~\eqref{eq:SMPframe} is equivalent to the permutation-equivariant $\mathcal{F}_{\text{GNN}^\prime}(\mathbf{A},\mathbf{F}; \mathbf{W}^\prime)$ if the parameters in $\mathcal{F}_{\text{output}}(\cdot)$ are all-zeros for $\tilde{E}$ and an identity matrix for $\mathbf{H}^{(L)}$.
\end{remark}
The result is straightforward from the definition.
\begin{corollary}\label{cor:powerful}
	For any task, Eq.~\eqref{eq:SMPframe} with a linear $\mathcal{F}_{\text{output}}(\cdot)$ in Remark~\ref{thm:permu} is at least as powerful as the permutation-equivariant $\mathcal{F}_{\text{GNN}^\prime}(\mathbf{A},\mathbf{F}; \mathbf{W}^\prime)$, i.e., the minimum training loss of using $\mathbf{H}$ in Eq.~\eqref{eq:SMPframe} is equal to or smaller than that using $ \mathbf{H}^{(L)} =\mathcal{F}_{\text{GNN}^\prime}(\mathbf{A},\mathbf{F}; \mathbf{W}^\prime)$.
\end{corollary}
In other words, SMP will not hinder the performance\footnote{Similar to previous analyses such as \cite{hamilton2017inductive,xu2019powerful}, we only consider the minimum training loss because the optimization landscapes and generalization gaps of deep neural networks are difficult to analyze analytically. We leave such explorations as future works.} even if the tasks are strictly permutation-equivariant, since the stochastic representations are concatenated with the permutation-equivariant GNNs followed by a linear mapping. In these cases, the linear SMP is equivalent to SGC~\cite{wu2019simplifying}.

Combining Theorem~\ref{thm:walksim2} and Corollary~\ref{cor:powerful}, the linear SMP instantiation in Eq.~\eqref{eq:SMP} is capable of handling both proximity-aware and permutation-equivariant tasks.

\subsection{Non-linear Extensions}\label{sec:nonlinearins}
One may be curious whether a more sophisticated variant of Eq.~\eqref{eq:SMPframe} can further improve the expressiveness of SMP. There are three adjustable components in Eq.~\eqref{eq:SMPframe}: two GNNs in propagating the stochastic matrix and node features, respectively, and an output function. In theory, adopting non-linear models as either component is able to enhance the expressiveness of SMP. Indeed, if we use a sufficiently expressive GNN in learning $\tilde{\mathbf{E}}$ instead of linear propagations, we can prove a more general version of Theorem~\ref{thm:walksim2}.

\begin{theorem}\label{thm:walksim3}
	
	For any length-$L$ walk-based proximity, i.e., $$\mathbf{S}_{i,j} = \mathcal{S}\left (\left\{v_i \leadsto v_j \right\} \right) = \mathcal{S}\left (\left\{v_i \leadsto v_j | \text{len}(v_i \leadsto v_j) \leq L \right\} \right),$$ where $\text{len}(\cdot)$ is the length of a walk, there exists an SMP variant in Eq.~\eqref{eq:SMPframe} with $\mathcal{F}_{\text{GNN}}\left(\mathbf{A},\mathbf{E}; \mathbf{W} \right)$ containing $L+1$ layers (including the input layer) to preserve that proximity if the following conditions hold: (1) The stochastic matrix $\mathbf{E}$ contains identifiable unique signals for different nodes, i.e. $\mathbf{E}_{i,:} \neq \mathbf{E}_{j,:},\forall i \neq j$. Here we assume that the Gaussian random vectors $\mathbf{E}$ are rounded to machine precision so that $\mathbf{E}$ is drawn from a countable subspace of $\;\mathbb{R}$. (2) The message-passing and updating functions in learning $\tilde{\mathbf{E}}$ are bijective. (3) The decoder function $\mathcal{F}_{\text{de}}(\cdot)$ also takes $\mathbf{E}$ as inputs and is universal approximation.

	\begin{proof}
		The proof of the theorem is given in Appendix~\ref{sec:sophi}.
	\end{proof}
\end{theorem}

We can also adopt more advanced methods for $\mathcal{F}_{\text{output}}(\cdot)$, such as attentions or even another GNN, so that the two GNNs are more properly integrated.

Although non-linear extensions of SMP can, in theory, increase the model expressiveness, they also take a higher risk of over-fitting due to model complexity, not to mention that the computational cost also increases. In practice, we find from our ablation studies that the linear SMP instantiation in Eq.~\eqref{eq:SMP} works reasonably well on most of the 
datasets (please refer to Section~\ref{sec:exp-ablation} for details).

\section{Experiments}\label{sec:exp}
In this section we report our extensive experimental studies. We first describe the experiment settings, benchmarks and baselines.  Then, we use a synthetic dataset to demonstrate the simultaneous needs of both permutation-equivariance and proximity-awareness in applications, illustrating the deficiencies of the existing GNNs in preserving the two properties and the capability of our SMP method.  We use a series of benchmark tasks, including link prediction, node classification, and pairwise node classification, to comprehensively examine the capability of our SMP method against the strong baselines.  Next, we conduct ablation studies of SMP.  Last, we evaluate the efficiency of our method.

\subsection{Experimental Setup}\label{sec:exp-setup}

Except for the proof-of-concept experiment in Section~\ref{sec:poc}, we use the following setup.

\subsubsection{Datasets}

We conduct experiments on the following \textbf{ten} datasets: two simulation datasets, \textbf{Grid} and \textbf{Communities} (\textbf{Comm} in abbreviation)~\cite{you2019position}, a communication dataset \textbf{Email}~\cite{you2019position}, two coauthor networks, \textbf{CS} and \textbf{Physics}~\cite{shchur2018pitfalls}, two protein interaction networks, \textbf{PPI}~\cite{hamilton2017inductive} and \textbf{PPA}~\cite{hu2020open}, and three benchmarks, \textbf{Cora}, \textbf{CiteSeer}, and \textbf{PubMed}~\cite{sen2008collective}. We summarize the statistics of datasets in Table~\ref{tab:dataset} and provide datasets details in Appendix~\ref{sec:datasets}.

These datasets cover a wide spectrum of application domains, various sizes, and with or without node features. Since Email and PPI contain more than one graph, we conduct experiments in an \emph{inductive setting}, i.e., the training, validation, and testing set are split with respect to different graphs. We repeat each experiment 5 times for all datasets except for PPA (3 times for each experiment on PPA), and report the averaged results and the standard deviations after the plus-minus signs.

\begin{table}[t]
	\centering
	\caption{The statistics of the datasets. For Email and PPI, \#Nodes and \#Edges are summed over all the graphs and the experiments are conducted in an inductive setting.}
	\begin{footnotesize}
		\begin{tabular}{@{}P{1.6cm}|P{0.85cm}P{1.05cm}P{1.05cm}P{0.95cm}P{1.1cm}@{}}
			\toprule
			Dataset          & \#Graphs & \#Nodes  & \#Edges     & \#Features & \#Classes         \\ \midrule
			Grid             & 1        & 400     & 760        & -         & -                \\
			Comm      & 1        & 400     & 3,800      & -         & 20               \\
			Email            & 7        & 1,005   & 25,571     & -         & 42               \\
			CS               & 1        & 18,333  & 81,894     & 6,805     & 15               \\
			Physics          & 1        & 34,493  & 247,962    & 8,415     & 5                \\
			PPI              & 24       & 56,944  & 818,716    & 50        & -              \\
			PPA              & 1        & 576,289 & 30,326,273 & 58        & -   \\
			Cora             & 1 & 2,708 & 5,429 & 1,433 & 7 \\
			CiteSeer         & 1 & 3,327 & 4,732 & 3,703 & 6 \\
			PubMed           & 1 & 19,717 & 44,338 & 500 & 3 \\ \bottomrule
		\end{tabular}
	\end{footnotesize}
	\label{tab:dataset}
\end{table}

\subsubsection{Baselines} We adopt two sets of baselines. The first set is permutation-equivariant GNNs including GCN~\cite{kipf2017semi}, GAT~\cite{velickovic2018graph}, and SGC~\cite{wu2019simplifying}. They are widely adopted GNN architectures. The second set contains P-GNN~\cite{you2019position}, a representative proximity-aware GNN. 

In comparing with the baselines, we mainly evaluate two variants of SMP with different $\mathcal{F}_{\text{output}}(\cdot)$: SMP-Identity, i.e., $\mathcal{F}_{\text{output}}(\cdot)$ as an identity mapping, and SMP-Linear, i.e., $\mathcal{F}_{\text{output}}(\cdot)$ as a linear function. Note that both variants adopt linear message-passing functions as SGC. We conduct ablation studies with more SMP variants in Section~\ref{sec:exp-ablation}.

For fair comparisons, we adopt the same architecture and hyper-parameters for all the methods (please refer to Appendix~\ref{sec:hyperp} for details). For datasets without node features, we adopt a constant vector as the node features.

\subsection{A Proof-of-concept Experiment}
\label{sec:poc}

\begin{table}
	\caption{The results of node classification measured in accuracy (\%) on the proof-of-concept synthetic dataset. The best result and the second-best result for each task, respectively, are in bold and underlined. }
	\centering
	\begin{tabular}{@{}l|l l l@{}}
		\toprule
		\multirow{2}{*}{Model}& \multicolumn{3}{c}{Node Labels}                              \\        
		& Community            &  Social Status         & Both         \\ \midrule
		Random                & 10.0                 & 50.0                   & 5.0          \\ \midrule
		SGC                   & 10.0$\pm$0.0         & 51.0$\pm$1.4           & 5.0$\pm$0.0  \\
		GCN                   & 8.7 $\pm$1.1         & {\ul91.6$\pm$1.8}      & {\ul8.9$\pm$0.9}   \\
		GAT                   & 10.0$\pm$0.0         & 73.4$\pm$12.9          & 5.0$\pm$0.0  \\ \midrule
		P-GNN                 & {\ul64.1$\pm$4.8}    & 54.9$\pm$9.8           & 5.6$\pm$1.2   \\ \midrule
		SMP-Linear            & \textbf{98.8$\pm$0.6}& \textbf{93.9$\pm$0.9}  & \textbf{93.8$\pm$1.6}  \\
		\bottomrule
	\end{tabular}
	\label{tab:exp-syn}
\end{table}

We first conduct a proof-of-concept experiment to demonstrate the importance of preserving both permutation-equivariance and proximity-awareness. We generate a synthetic dataset similar to the intuition behind the example in Figure~\ref{fig:example} as follows. First, the nodes are randomly partitioned into a set of communities. The nodes within the same community have a higher probability of forming edges than the nodes in different communities, i.e., the well-known stochastic block model~\cite{newman2018networks}. Then, within each community, we generate a social status for each node with two possible choices. If a node is \emph{active}, it has a high probability of forming edges with other nodes in the same community. Otherwise, the node has a low probability of forming edges with others, i.e., \emph{inactive}. From the above generating process, we can see that proximity-awareness is essential to predict which community a node belongs to, since nodes within the same community have large proximities. To predict whether a node is active or inactive, permutation-equivariance is helpful, since the social status serves as a type of centrality measurements. Please refer to Appendix~\ref{sec:datasets} for further details of the synthetic dataset. 

\begin{table*}[t]
	\begin{minipage}{0.67\linewidth}\centering
		\caption{The results of link prediction tasks measured in AUC (\%). The best result and the second-best result for each dataset, respectively, are in bold and underlined.}
		\begin{tabular}{@{}l|llllll@{}}
			\toprule
			Model          & Grid     &  Comm           & Email         & CS            & Physics            & PPI         \\ \midrule
			SGC                            & 57.6$\pm$3.8          & 51.9$\pm$1.6          & 68.5$\pm$7.0          & {\ul 96.5$\pm$0.1}    & \textbf{96.6$\pm$0.1} & 80.5$\pm$0.4          \\
			GCN                            & 61.8$\pm$3.6          & 50.3$\pm$2.5          & 67.4$\pm$6.9          & 93.4$\pm$0.3          & 93.8$\pm$0.2          & 78.0$\pm$0.4          \\
			GAT                            & 61.0$\pm$5.5          & 51.1$\pm$1.6          & 53.5$\pm$6.3          & 93.7$\pm$0.9          & 94.1$\pm$0.4          & 79.3$\pm$0.5          \\ \midrule
			P-GNN                 & {\ul 73.4$\pm$6.0}    & {\ul 97.8$\pm$0.6}    & 70.9$\pm$6.4          & 82.2$\pm$0.5          & Out of memory         & 80.8$\pm$0.4          \\ \midrule
			SMP-Identity                   & 55.1$\pm$4.8          & \textbf{98.0$\pm$0.7} & {\ul 72.9$\pm$5.1}    & {\ul 96.5$\pm$0.1}    & {\ul 96.5$\pm$0.1}    & {\ul 81.0$\pm$0.2}    \\
			SMP-Linear                     & \textbf{73.6$\pm$6.2} & 97.7$\pm$0.5          & \textbf{75.7$\pm$5.0} & \textbf{96.7$\pm$0.1} & 96.1$\pm$0.1          & \textbf{81.9$\pm$0.3} \\
			\bottomrule
		\end{tabular}
		\label{tab:exp-link}
	\end{minipage}
	\hspace{0.25cm}
	\begin{minipage}{0.3\linewidth}\centering
		\caption{The results of link prediction on the PPA dataset. }
		\begin{tabular}{@{}l|l@{}}
			\toprule
			Model                & Hits@100            \\ \midrule
			SGC                  & 0.1187$\pm$0.0012   \\
			GCN                  & 0.1867$\pm$0.0132   \\
			GraphSAGE            & 0.1655$\pm$0.0240 \\ \midrule
			P-GNN                & Out of Memory        \\
			\midrule
			Node2vec             & 0.2226$\pm$0.0083 \\
			Matrix Factorization & \underline{0.3229$\pm$0.0094} \\ \midrule
			SMP-Identity	& 0.2018$\pm$0.0148 \\
			SMP-Linear       & \textbf{0.3582$\pm$0.0070} \\
			\bottomrule
		\end{tabular}
		\label{tab:exp-ppa}
	\end{minipage}
\end{table*}

We conduct experiments on the synthetic dataset for the node classification task, i.e., predicting the node labels. We consider the following three cases. (1) \textbf{Community}: The node label is the community that the node belongs to. (2) \textbf{Social Status}: The node label is the social status of the node. (3) \textbf{Both}: The node label is the Cartesian product of (1) and (2), i.e., every community and social status pair is a distinct label. We use a softmax layer on the learned node representations as the classifier, and use accuracy, i.e., the percentage of nodes correctly classified, as the evaluation metric. We omit the results of SMP-Identity since the node representations in SMP-Identity have a fixed dimensionality that does not match the number of classes.

Table~\ref{tab:exp-syn} shows the results, which are consistent with our analyses. The permutation-equivariant GNNs perform reasonably well on predicting the social status labels but cannot discover communities, since node proximities are not preserved in those methods. P-GNN manages to handle community labels well, but performs poorly for social status labels. None of them can handle the most challenging setting where both properties are needed to predict the node labels of community and status. 

SMP performs consistently well in all three cases. The results clearly show that SMP can simultaneously preserve permutation-equivariance and proximity-awareness when needed and retain highly competitive performance for each property. In fact, for the community labels, SMP significantly outperforms P-GNN, demonstrating that SMP can better preserve proximities between nodes.

Next, we report experimental results on benchmark datasets.

\subsection{Link Prediction}\label{sec:exp-linkprediction}
Link prediction predicts missing links in a graph. We randomly split the edges into three exclusive parts of relative sizes 80\%, 10\% and 10\%, and use them for training, validation, and testing, respectively. Besides these positive samples, we obtain negative samples by randomly sampling an equal number of node pairs that do not have edges for training/validation/testing. For all the methods, we set a simple classifier: $\operatorname{Sigmoid}(\mathbf{H}_i^T \mathbf{H}_j)$, i.e., use the inner product to predict whether a node pair $(v_i, v_j)$ forms a link, and use AUC (area under the ROC curve) as the evaluation metric. One exception to this setting is that on the PPA dataset, we follow the splits and evaluation metric (i.e., Hits@100) provided by the dataset~\cite{hu2020open}. Limited by space, the results for three benchmarks (Cora, CiteSeer, and PubMed) are shown in Appendix~\ref{sec:exp_addresults_LP}.

The results except PPA are shown in Table~\ref{tab:exp-link}. SMP achieves the best results on five out of the six datasets and is highly competitive (the second-best result) on the other (Physics). The results demonstrate the effectiveness of SMP in link prediction tasks. We attribute the strong performance of SMP to its capability of maintaining both proximity-awareness and permutation-equivariance properties.

On Grid, Communities, Email, and PPI, both SMP and P-GNN outperform the permutation-equivariant GNNs, confirming the importance of preserving node proximities. Although SMP is simpler and more efficient than P-GNN, SMP reports even better results.

When node features are available (CS, Physics, and PPI), SGC outperforms GCN and GAT. The results re-validate the findings in SGC~\cite{wu2019simplifying} and LightGCN~\cite{he2020light} 
that the non-linearity in GNNs is not necessarily indispensable. Some plausible reasons include that the additional model complexity brought by non-linear operators makes the models tend to overfit or difficult to be trained.
On those datasets, SMP retains comparable performance on two coauthor graphs and shows better performance on PPI, possibly because the node features on the protein graphs are less informative than the node features on coauthor graphs for predicting links. Thus, preserving graph structure is more beneficial on PPI.  As we experiment on Email and PPI in the inductive setting, the results show that SMP also can handle inductive tasks well.

The results on PPA are shown in Table~\ref{tab:exp-ppa}. SMP outperforms all the baselines, showing that it can handle large-scale graphs with millions of nodes and edges. PPA is part of a recently released Open Graph Benchmark (OGB)~\cite{hu2020open}. The superior performance on PPA further demonstrates the effectiveness of SMP in link prediction.

\subsection{Node Classification}\label{sec:exp-nodeclassification}

In the task of node classification, we need ground-truths in the evaluation. Thus, we only adopt datasets with node labels. Specifically, for CS and Physics, we adopt 20/30 labeled nodes per class for training/validation and the rest for testing~\cite{shchur2018pitfalls}. For Comm, we adjust the number as 5/5/10 labeled nodes per class for training/validation/testing.
For Cora, CiteSeer, and PubMed, we use the default splits that come with the datasets.
We do not adopt Email because some graphs in the dataset are too small to show stable results. We also exclude PPI since it is a multi-label dataset. Other settings are the same as Section~\ref{sec:poc}.

\begin{table}[t]
	\centering
	\caption{The results of node classification tasks measured by accuracy (\%). The best results and the second-best results for each dataset, respectively, are in bold and underlined. OOM represents out of memory.}
	\begin{small}
		\begin{adjustbox}{max width=0.49\textwidth}
			\begin{tabular}{@{}c|cccccc@{}}
				
				\toprule
				Model          & Comm      & CS            & Physics                                           & Cora & CiteSeer & PubMed          \\ \midrule
				SGC                            & 7.1$\pm$2.1             & 67.2$\pm$12.8         & 92.3$\pm$1.6          & 76.9$\pm$0.2          & 63.6$\pm$0.0          & 74.2$\pm$0.1          \\
				GCN                            & {\ul 7.5$\pm$1.2}       & {\ul 91.1$\pm$0.7}    & {\ul 93.1$\pm$0.8}    & {\ul 81.4$\pm$0.5}    & \textbf{71.3$\pm$0.5} & \textbf{79.3$\pm$0.4} \\
				GAT                            & 5.0$\pm$0.0             & 90.5$\pm$0.5          & \textbf{93.1$\pm$0.4} & \textbf{82.9$\pm$0.5} & {\ul 71.2$\pm$0.6}    & {\ul 77.9$\pm$0.5}    \\ \midrule
				P-GNN                           & 5.2$\pm$0.5             & 77.6$\pm$7.6          & OOM         & 59.2$\pm$1.5          & 55.7$\pm$0.9          & OOM         \\ \midrule
				SMP-Linear                     & \textbf{99.9$\pm$0.3}   & \textbf{91.5$\pm$0.8} & {\ul 93.1$\pm$0.8}    & 80.9$\pm$0.8          & 68.2$\pm$1.0          & 76.5$\pm$0.8          \\ \bottomrule         
			\end{tabular}
		\end{adjustbox}
	\end{small}
	\label{tab:exp-node}
\end{table}

The results are shown in Table~\ref{tab:exp-node}.
SMP reports nearly perfect results on Comm. Since the node labels are generated by graph structures on Comm and there are no node features, a model has to be proximity-aware to handle Comm well. P-GNN, which shows promising results in the link prediction task, fails miserably here.

\begin{table*}[t]
	\centering
	\caption{The results of pairwise node classification tasks measured in AUC (\%). The best result and the second-best result for each dataset, respectively, are in bold and underlined.}
	\begin{tabular}{@{}l|lllllll@{}}
		\toprule
		Model          & Comm             & Email                      & CS                    & Physics  & Cora & CiteSeer & PubMed            \\ \midrule
		SGC                            & 67.4$\pm$2.4             & 56.3$\pm$5.4          & \textbf{99.8$\pm$0.0} & {\ul 99.6$\pm$0.0}    & {\ul 99.2$\pm$0.3}    & \textbf{95.5$\pm$0.7} & 92.3$\pm$0.3          \\
		GCN                            & 64.9$\pm$2.3             & 55.0$\pm$5.7          & 96.8$\pm$0.7          & \textbf{99.7$\pm$0.1} & 97.7$\pm$0.6          & 92.9$\pm$1.2          & \textbf{94.8$\pm$0.4} \\
		GAT                            & 52.5$\pm$1.3             & 47.7$\pm$2.7          & 95.2$\pm$0.6          & 96.3$\pm$0.2          & 91.6$\pm$0.7          & 73.6$\pm$2.7          & 87.1$\pm$0.2          \\ \midrule
		P-GNN                           & \underline{98.6$\pm$0.5} & {\ul 63.3$\pm$5.5}    & 90.0$\pm$0.5          & Out of memory         & 85.5$\pm$1.2          & 49.8$\pm$1.8          & Out of memory         \\ \midrule
		SMP-Identity                   & \textbf{98.8$\pm$0.5}    & 56.9$\pm$4.1          & \underline{99.7$\pm$0.0}          & {\ul 99.6$\pm$0.0}    & {\ul 99.2$\pm$0.2}    & 95.2$\pm$1.1          & 91.9$\pm$0.3          \\
		SMP-Linear                     & \textbf{98.8$\pm$0.5}    & \textbf{74.5$\pm$4.1} & \textbf{99.8$\pm$0.0} & {\ul 99.6$\pm$0.0}    & \textbf{99.3$\pm$0.3} & {\ul 95.3$\pm$0.4}    & {\ul 93.4$\pm$0.2}
		\\
		\bottomrule
	\end{tabular}
	\label{tab:exp-pair}
\end{table*}

\begin{table*}[ht]
	\centering
	\caption{The results of graph reconstruction measured in AUC (\%). The best and the second-best results for each dataset, respectively, are in bold and underlined.}
	\begin{tabular}{{@{}c| c c c c c c c c c@{}}}
		\toprule
		Model                 & Grid            &  Comm     & Email             & CS                & Physics          & PPI              & Cora            &  CiteSeer       & PubMed\\ \midrule
		SGC                   & 74.8$\pm$0.4    & 65.4$\pm$1.6     & 71.6$\pm$0.3      & 66.7$\pm$0.1      & 66.2$\pm$0.0     & 76.3$\pm$0.2     & 56.7$\pm$9.7    & 58.5$\pm$0.1    & 71.9$\pm$0.1 \\
		GCN                   & 73.0$\pm$0.3    & 63.7$\pm$1.2     & 72.5$\pm$0.4      & 75.5$\pm$0.4      &\as{76.8$\pm$0.4} & 79.2$\pm$0.4     & 68.2$\pm$3.9    & 69.8$\pm$8.0    &\as{77.2$\pm$2.1} \\
		GAT                   & 59.6$\pm$1.2    & 52.9$\pm$1.1     & 56.9$\pm$1.9      & 57.0$\pm$1.4      & 59.1$\pm$0.7     & 61.1$\pm$1.9     & 57.8$\pm$1.0    & 63.2$\pm$1.5    & 58.8$\pm$0.8 \\ \midrule
		P-GNN                  &\ab{99.4$\pm$0.1}&\as{97.7$\pm$0.1} &\as{85.6$\pm$0.8}  &\ab{97.2$\pm$0.6}  & Out of memory   &\as{85.2$\pm$0.6} &\ab{98.1$\pm$0.6}&\ab{99.7$\pm$0.1}& Out of memory          \\ \midrule
		SMP-Identity          &\as{99.2$\pm$0.1}&97.5$\pm$0.1      & 80.0$\pm$0.3      & 77.1$\pm$2.3      & 73.7$\pm$0.3     & 79.5$\pm$0.2     & 89.7$\pm$5.7    & 97.1$\pm$0.8    & 77.0$\pm$0.1 \\
		SMP-Linear            & 99.1$\pm$0.1    &\ab{97.8$\pm$0.1} &\ab{86.7$\pm$0.2}  &\as{96.3$\pm$0.2}  &\ab{95.5$\pm$0.2} &\ab{85.5$\pm$0.1} &\as{96.3$\pm$0.1} &\as{98.2$\pm$0.1}&\ab{95.8$\pm$0.2}\\ \bottomrule
	\end{tabular}
	\label{tab:network-reconstruction}
\end{table*}

On the other five graphs, SMP reports highly competitive performance. These graphs are commonly-used benchmarks for GNNs.  P-GNN, which completely ignores permutation-equivariance, performs poorly as expected. In contrast, SMP manages to be competitive with the permutation-equivariant GNNs, as endorsed by Remark~\ref{thm:permu}. In fact, SMP even shows better results than its counterpart, SGC, indicating that preserving proximities is also helpful.

\subsection{Pairwise Node Classification}\label{sec:exp_addresults_NC}
We follow P-GNN~\cite{you2019position} and experiment on pairwise node classification, i.e., predicting whether two nodes have the same label.
Compared with node classification in Section~\ref{sec:exp-nodeclassification}, pairwise node classification focuses more on the relation between nodes and thus more likely to require a model to be proximity-aware.

Similar to link prediction, we split the positive samples (i.e., node pairs with the same label) into an 80\%-10\%-10\% training-validation-testing set with an equal number of randomly sampled negative pairs. For large graphs, since the possible positive samples are intractable (i.e. $O(N^2)$), we use a random subset. As we also need node labels as the ground-truth, we only conduct pairwise node classification on the datasets when node labels are available. We also exclude the results on PPI since the dataset is multi-labeled and cannot be used in a pairwise setting~\cite{you2019position}. Similar to the link prediction task in Section~\ref{sec:exp-linkprediction}, we adopt a simple inner product classifier and use AUC as the evaluation metric.

The results are shown in Table~\ref{tab:exp-pair}. We observe consistent results as the link prediction task, i.e., SMP reports the best results on four datasets and the second-best results on the other three datasets. These results again verify that SMP can effectively preserve and utilize node proximities while retaining comparable performance when the tasks are more permutation-equivariant like, e.g., on CS, Physics, and the three benchmarks (Cora, CiteSeer, and PubMed).

\subsection{Graph Reconstruction}\label{sec:exp_addresults_nr}

To examine whether SMP can indeed preserve node proximities, we conduct experiments on graph reconstruction~\cite{wang2016structural}, i.e., using the node representations learned by GNNs to reconstruct the edges of the graph. Graph reconstruction corresponds to the first-order proximity between nodes, i.e., whether two nodes directly have a connection, which is the most straightforward node proximity~\cite{tang2015line}. Specifically, following link prediction and pairwise node classification, we adopt the inner product classifier $\operatorname{Sigmoid}(\mathbf{H}_i^T \mathbf{H}_j)$ and use AUC as the evaluation metric. To control the impact of node features (i.e., many graphs exhibit assortative mixing~\cite{newman2018networks}, thus even models only using node features can reconstruct the edges to a certain extent), we do not use node features for all the models.

The results are reported in Table~\ref{tab:network-reconstruction}. The results show that SMP greatly outperforms permutation-equivariant GNNs such as GCN and GAT for the graph reconstruction task, clearly demonstrating that SMP can better preserve node proximities. P-GNN shows highly competitive results as SMP. However, similar to the other tasks, the intensive memory usage makes P-GNN unable to handle medium-scale graphs such as Physics and PubMed.

\begin{table*}[ht]
	\centering
	\caption{The ablation study of SMP variants for the link prediction task. Datasets except PPA are measured by AUC (\%) and PPA is measured by Hits@100. The best result and the second-best result for each dataset are in bold and underlined, respectively.}
	\begin{tabular}{@{}l|llllll|l@{}}
		\toprule
		Model          & Grid                      & Comm          & Email                   & CS              & Physics         & PPI                  & PPA                  \\ \midrule
		SMP-Identity                   & 55.1$\pm$4.8          & \textbf{98.0$\pm$0.7}    & 72.9$\pm$5.1          & {\ul 96.5$\pm$0.1}    & \textbf{96.5$\pm$0.1} & 81.0$\pm$0.2          & 0.2018$\pm$0.0148             \\
		SMP-Linear                     & {\ul 73.6$\pm$6.2}    & 97.7$\pm$0.5          & \textbf{75.7$\pm$5.0} & \textbf{96.7$\pm$0.1} & {\ul 96.1$\pm$0.1}    & {\ul 81.9$\pm$0.3}    & \underline{0.3582$\pm$0.0070} \\
		SMP-MLP                        & 72.1$\pm$4.3          & {\ul 97.8$\pm$0.6}          & 62.7$\pm$8.1          & 88.9$\pm$0.8          & 89.2$\pm$0.4          & 80.1$\pm$0.3          & 0.2035$\pm$0.0038             \\
		SMP-Linear-GCN$_{\text{feat}}$ & 72.8$\pm$4.2          & \textbf{98.0$\pm$0.4} & {\ul 74.2$\pm$3.9}    & 92.9$\pm$0.6          & 94.3$\pm$0.2          & \textbf{82.3$\pm$1.0} & \textbf{0.4090$\pm$0.0087}    \\
		SMP-Linear-GCN$_{\text{both}}$ & \textbf{80.5$\pm$3.9} & 97.3$\pm$0.7          & 73.4$\pm$5.5          & 89.8$\pm$2.0          & 91.7$\pm$0.2          & 79.7$\pm$0.3          & 0.2125$\pm$0.0232            \\
		\bottomrule
	\end{tabular}
	\label{tab:exp-ablation}
\end{table*}

\begin{table*}[t]
	\centering
	\caption{The results of comparing whether the stochastic signals $\mathbf{E}$ are fixed or not during different training epochs for the link prediction task. The better of the two results are in bold.}
	\begin{tabular}{@{}c|l|llll|ll@{}}
		\toprule
		Model                       &  $\mathbf{E}$ & Grid                  &  Comm          & CS                    & Physics               & Email                  & PPI                   \\      \midrule
		\multirow{2}{*}{SMP-Identity}&  Fixed        & 55.1$\pm$4.8          & \textbf{98.0$\pm$0.7} & \textbf{96.5$\pm$0.1} & 96.5$\pm$0.1          & \textbf{75.9$\pm$3.9}  & 80.4$\pm$0.4          \\
		&  Not Fixed    & \textbf{55.2$\pm$4.1} & 97.6$\pm$0.7          & 96.4$\pm$0.1          & 96.5$\pm$0.1          & 72.9$\pm$5.1           & \textbf{81.0$\pm$0.2} \\ \midrule
		\multirow{2}{*}{SMP-Linear}  &  Fixed        & \textbf{73.6$\pm$6.2} & \textbf{97.7$\pm$0.5} & \textbf{96.7$\pm$0.1} & 96.1$\pm$0.1          & 71.3$\pm$3.9           & 71.5$\pm$0.7          \\
		&  Not Fixed    & 64.4$\pm$2.9          & 97.4$\pm$0.1          & 96.2$\pm$0.1          & 96.1$\pm$0.1          & \textbf{75.7$\pm$5.0}  & \textbf{81.9$\pm$0.3} \\ \bottomrule
	\end{tabular}
	\label{tab:exp-ab-fix}
\end{table*}

\begin{table}[t]
	\centering
	\caption{The average running time (milliseconds) for each epoch (including both training and testing), on the link prediction task. OOM represents out of memory.}
	\begin{small}
		\begin{adjustbox}{max width=0.49\textwidth}
			\begin{tabular}{@{}l|llllll@{}}
				\toprule
				Model          & Grid & Comm & Email & CS  & Physics & PPI \\ \midrule
				SGC                            & 25           & 28                  & 58            & 210        & 651             & 704         \\
				GCN                            & 25           & 35                  & 75            & 214        & 612             & 784         \\
				GAT                            & 36           & 43                  & 140           & 258        & 801             & 919         \\ \midrule
				P-GNN                           & 81           & 84                  & 206           & 19,340     & OOM             & 6,521       \\ \midrule
				SMP-Identity                   & 26           & 37                  & 96            & 284        & 751             & 840         \\
				SMP-Linear                     & 28           & 26                  & 84            & 212        & 616             & 832         \\
				SMP-MLP                        & 23           & 28                  & 83            & 237        & 614             & 831         \\
				SMP-Linear-GCN$_{\text{feat}}$ & 23           & 29                  & 90            & 231        & 636             & 855         \\
				SMP-Linear-GCN$_{\text{both}}$ & 34           & 40                  & 95            & 228        & 626             & 895         \\
				\bottomrule
			\end{tabular}
		\end{adjustbox}
	\end{small}
	\label{tab:exp-time}
\end{table}

\subsection{Ablation Studies}\label{sec:exp-ablation}
We conduct ablation studies by comparing different SMP variants, including SMP-Identity, SMP-Linear, and additional three variants. 
\begin{itemize}[leftmargin=0.5cm]
	\item In SMP-MLP, we set $\mathcal{F}_{\text{output}}(\cdot)$ to a fully-connected network with one hidden layer. 
	\item In SMP-Linear-GCN$_{\text{feat}}$, we set  $\mathcal{F}_{\text{GNN}^\prime}(\mathbf{A},\mathbf{F}; \mathbf{W}^\prime)$ in Eq.~\eqref{eq:SMPframe} to a GCN~\cite{kipf2017semi}, i.e., induce non-linearity in the message-passing for features. $\mathcal{F}_{\text{GNN}}\left(\mathbf{A},\mathbf{E}; \mathbf{W} \right)$ and $\mathcal{F}_{\text{output}}(\cdot)$ are still linear.
	\item   In SMP-Linear-GCN$_{\text{both}}$, we set both GNNs in Eq.~\eqref{eq:SMPframe}, i.e., $\mathcal{F}_{\text{GNN}}\left(\mathbf{A},\mathbf{E}; \mathbf{W} \right)$ and $\mathcal{F}_{\text{GNN}^\prime}(\mathbf{A},\mathbf{F}; \mathbf{W}^\prime)$, to a GCN~\cite{kipf2017semi}, i.e., induce non-linearity in message-passing for both features and stochastic representations. $\mathcal{F}_{\text{output}}(\cdot)$ is linear.
\end{itemize}
We show the results of link prediction in Table~\ref{tab:exp-ablation}. The results for node classification and pairwise node classification, which show similar conclusions, are provided in Appendix~\ref{sec:exp_addresults_ablation}.

In general, SMP-Linear shows impressive performance, achieving the best or second-best results on six datasets and highly competitive on the other (Comm). SMP-Identity, which does not have learnable parameters in the output function, performs slightly worse. The results demonstrate the importance of adopting a learnable linear layer in the output function, which is consistent with Remark~\ref{thm:permu}. SMP-MLP does not lift the performance in general, showing that adding extra complexities in $\mathcal{F}_{\text{output}}(\cdot)$ brings no gain in those datasets.
SMP-Linear-GCN$_{\text{feat}}$ reports the best results on Communities, PPI, and PPA, indicating that adding extra non-linearities in propagating node features is helpful for some graphs. 
SMP-Linear-GCN$_{\text{both}}$ reports the best results on Gird with a considerable margin. Recall that Grid has no node features. The results indicate that inducing non-linearities can help the stochastic representations to better capture proximities for some graphs.

We also assess the effects of whether the stochastic signals $\mathbf{E}$ are fixed or not during different training epochs for our proposed SMP. For brevity, we only report the results of link prediction in Table~\ref{tab:exp-ab-fix}. The results show that fixing $\mathbf{E}$ usually leads to better results on transductive datasets (recall that datasets except Email and PPI are transductive) and resampling $\mathbf{E}$ leads to better results on inductive datasets. The results are consistent with our analysis in Section~\ref{sec:SMP}.

\subsection{Efficiency}\label{sec:runtime}
To compare the efficiency of different methods quantitatively, we report the running time of different methods in Table \ref{tab:exp-time}.
The results are averaged over 3,000 epochs on an NVIDIA TESLA M40 GPU with 12 GB of memory.

The results show that SMP is computationally efficient, i.e., only marginally slower than SGC and comparable to GCN. P-GNN is at least an order of magnitude slower except for the extremely small graphs such as Grid, Communities, or Email, which have no more than a thousand nodes. In addition, the expensive memory cost makes P-GNN unable to work on large-scale graphs.

\section{Conclusion}\label{sec:con}

In this paper, we propose SMP, a general and simple GNN to preserve both proximity-awareness and permutation-equivariance. We augment the existing GNNs with stochastic node representations. We prove that SMP can enable GNNs to preserve node proximities and is equivariant to a permutation-equivariant GNN with certain parametrization. Our experimental results demonstrate the effectiveness and efficiency of SMP.

\ifCLASSOPTIONcompsoc
\section*{Acknowledgments}

\else
\section*{Acknowledgment}
\fi

This work was supported in part by National Key Research and Development Program of China (No. 2020AAA0106300), National Natural Science Foundation of China (No. U1936219, No. 62141607, No. 62050110), and Beijing Academy of Artificial Intelligence (BAAI). All opinions, findings, conclusions and recommendations in this paper are those of the authors and do not necessarily reflect the views of the funding agencies. Peng Cui and Wenwu Zhu are corresponding authors.

\bibliographystyle{IEEEtran}
\bibliography{sample-sigconf}


%




\ifCLASSOPTIONcaptionsoff
  \newpage
\fi

\begin{IEEEbiography}[{\includegraphics[width=1in,height=1.3in,clip,keepaspectratio]{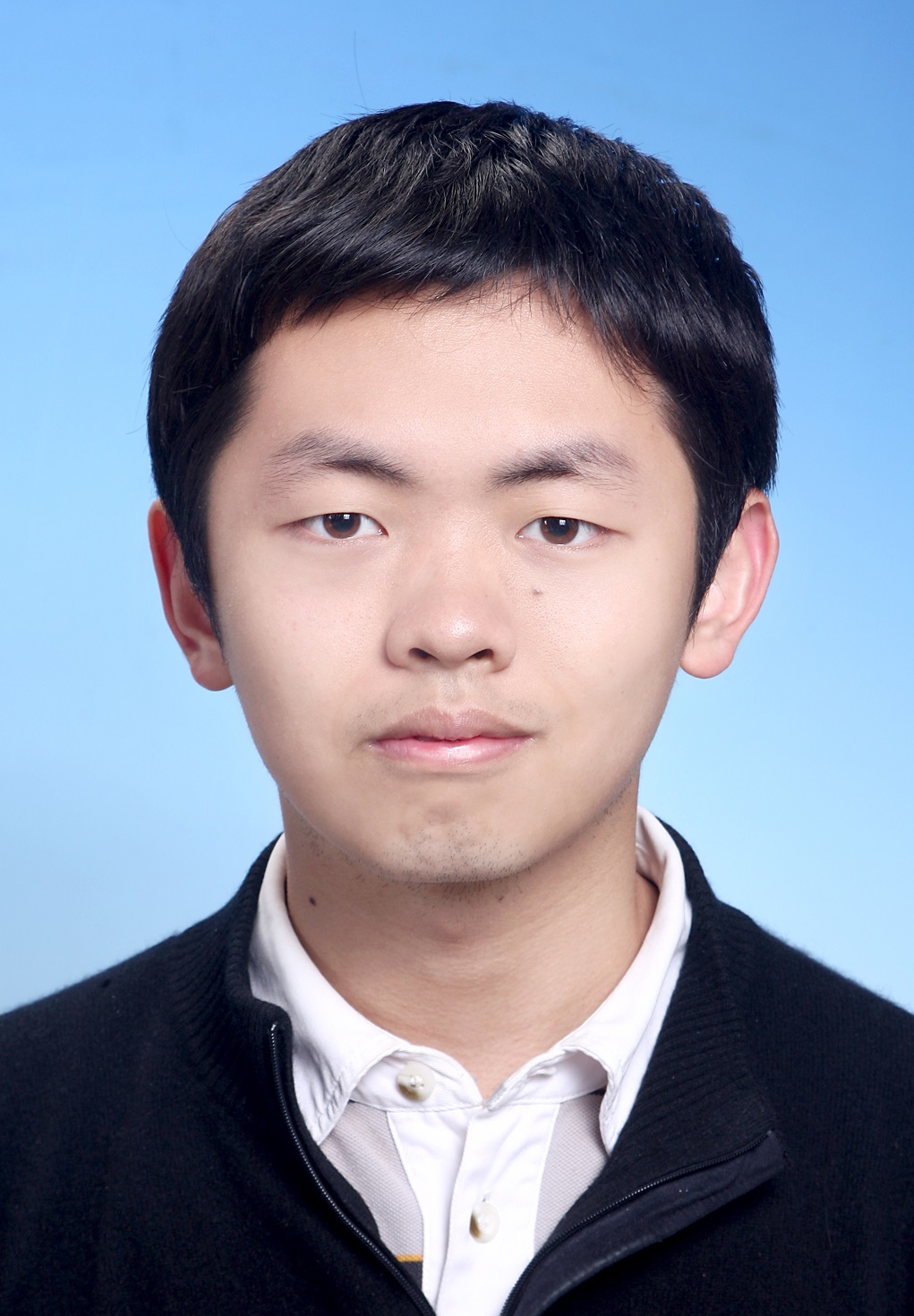}}]{Ziwei Zhang} received his Ph.D. from the Department of Computer Science and Technology, Tsinghua University, in 2021. He is currently a postdoc researcher in the Department of Computer Science and Technology at Tsinghua University. His research interests focus on machine learning on graphs, including graph neural network (GNN) and network representation learning. He has published over a dozen papers in prestigious conferences and journals, including KDD, AAAI, IJCAI, NeurIPS, and TKDE.
\end{IEEEbiography}

\begin{IEEEbiography}[{\includegraphics[width=1in,height=1.25in,clip,keepaspectratio]{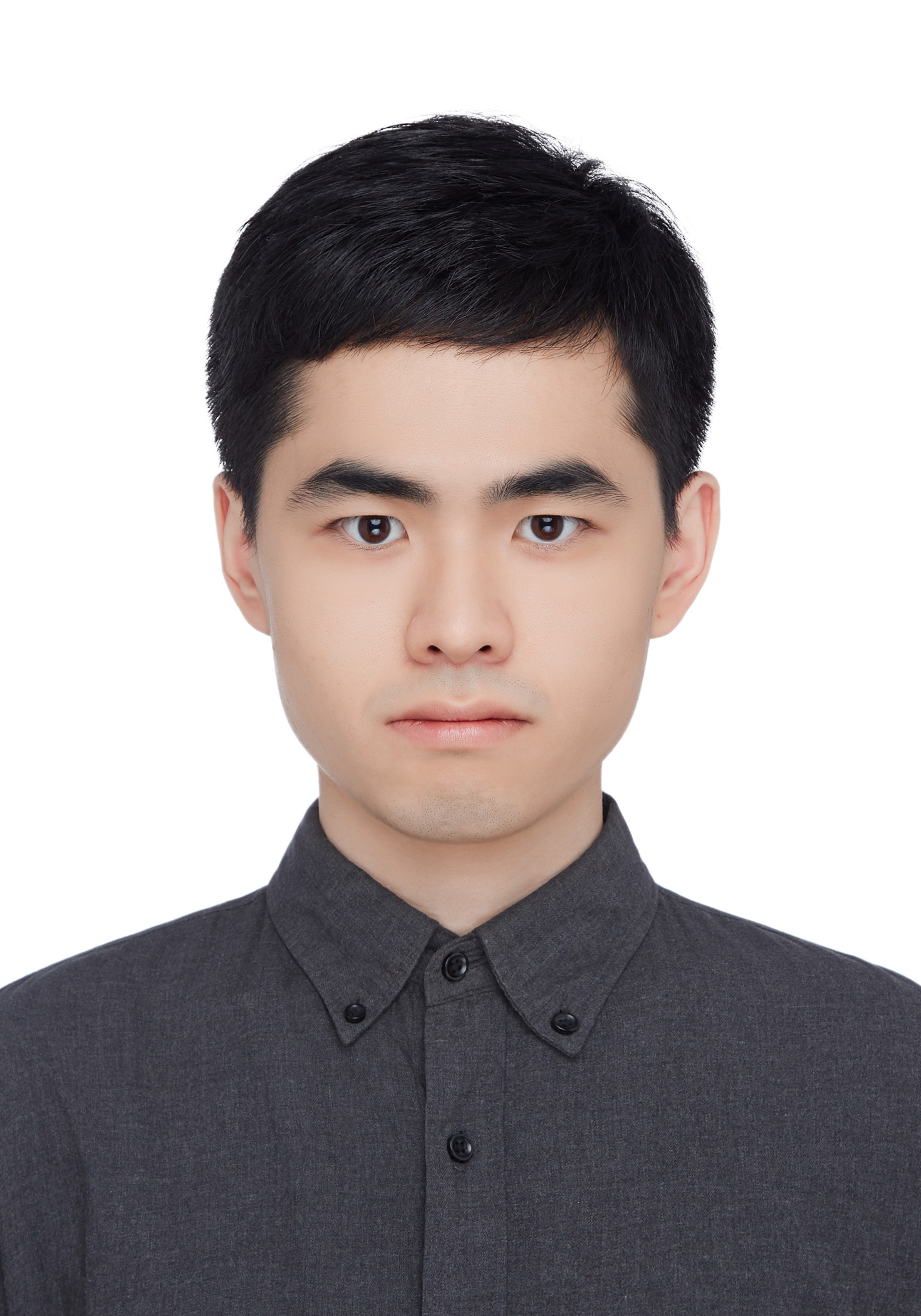}}]{Chenhao Niu} received his B.E. degree in Computer Science and Technology from Tsinghua University, 2020. He is currently a M.Sc. student in Machine Learning at Carnegie Mellon University. His research interests are mainly in machine learning with graph neural networks and generative models, and he has related publications on KDD, AISTATS and TKDE.
\end{IEEEbiography}

\begin{IEEEbiography}[{\includegraphics[width=1in,height=1.3in,clip,keepaspectratio]{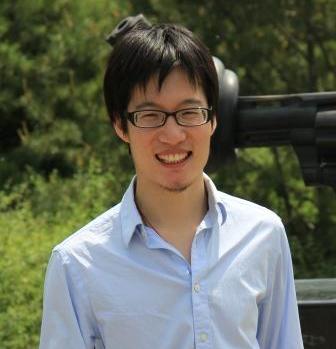}}]{Peng Cui} is an Associate Professor with tenure in Tsinghua University. He got his PhD degree from Tsinghua University in 2010. His research interests include causally-regularized machine learning, network representation learning, and social dynamics modeling. He has published more than 100 papers in prestigious conferences and journals in data mining and multimedia. His recent research won the IEEE Multimedia Best Department Paper Award, SIGKDD 2016 Best Paper Finalist, ICDM 2015 Best Student Paper Award, SIGKDD 2014 Best Paper Finalist, IEEE ICME 2014 Best Paper Award, ACM MM12 Grand Challenge Multimodal Award, and MMM13 Best Paper Award. He is PC co-chair of CIKM2019 and MMM2020, SPC or area chair of WWW, ACM Multimedia, IJCAI, AAAI, etc., and Associate Editors of IEEE TKDE, IEEE TBD, ACM TIST, and ACM TOMM etc. He received ACM China Rising Star Award in 2015, and CCF-IEEE CS Young Scientist Award in 2018. He is now a Distinguished Member of ACM and CCF, and a Senior Member of IEEE.
\end{IEEEbiography}

\begin{IEEEbiography}[{\includegraphics[width=1in,height=1.3in,clip,keepaspectratio]{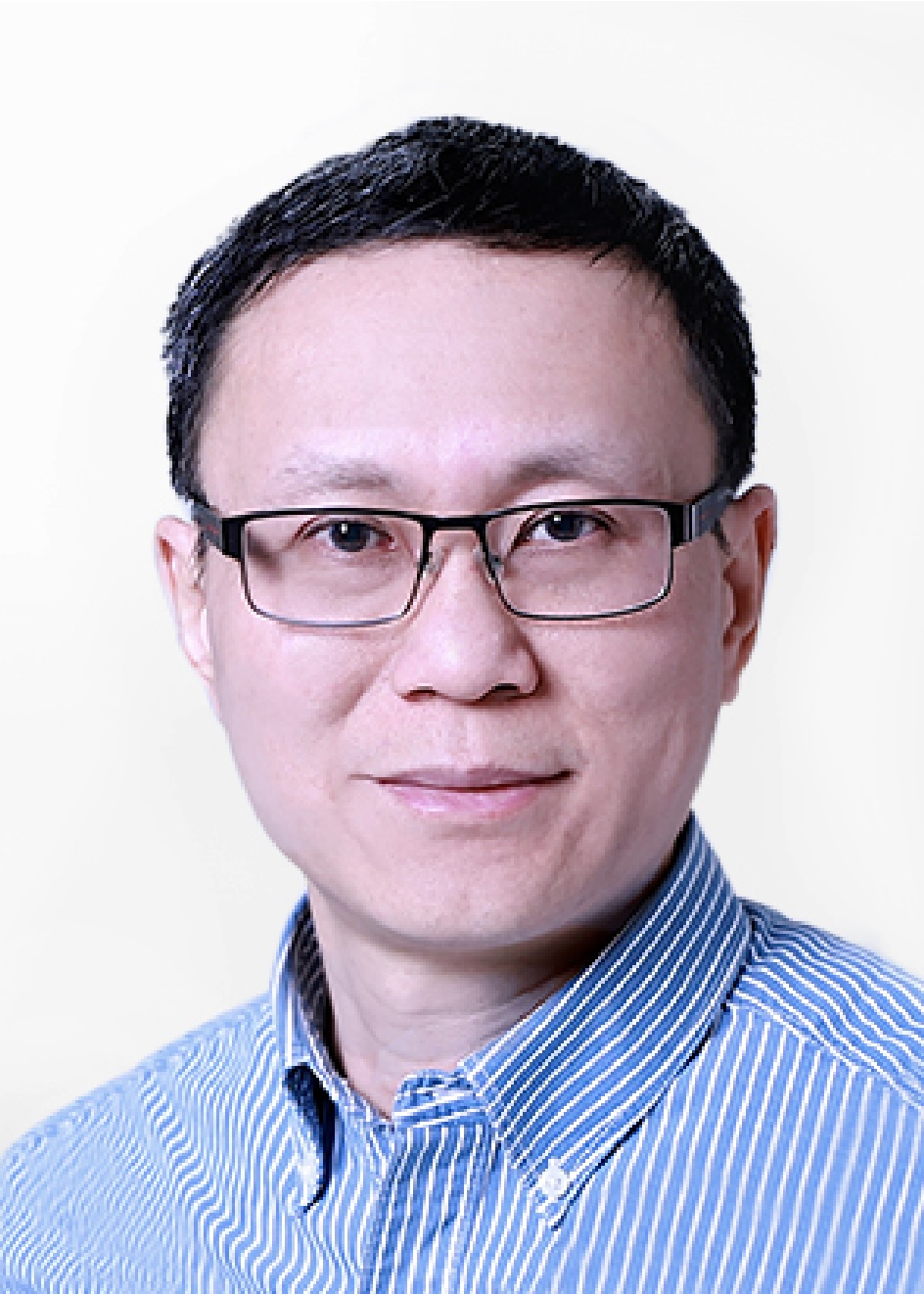}}]{Jian Pei} is a Professor in the School of Computing Science at Simon Fraser University. He is a renown leading researcher in the general areas of data science, big data, data mining, and database systems. He is recognized as a Fellow of the Royal Society of Canada (Canada's national academy), the Canadian Academy of Engineering, ACM and IEEE. He is one of the most cited authors in data mining, database systems, and information retrieval. Since 2000, he has published one textbook, two monographs and over 300 research papers in refereed journals and conferences, which have been cited extensively by others. His research has generated remarkable impact substantially beyond academia. For example, his algorithms have been adopted by industry in production and popular open source software suites. Jian Pei also demonstrated outstanding professional leadership in many academic organizations and activities. He was the editor-in-chief of the IEEE Transactions of Knowledge and Data Engineering (TKDE) in 2013-16, the chair of ACM SIGKDD in 2017-2021, and a general co-chair or program committee co-chair of many premier conferences.  He maintains a wide spectrum of industry relations with both global and local industry partners. He is an active consultant and coach for industry.  He received many prestigious awards, including the 2017 ACM SIGKDD Innovation Award, the 2015 ACM SIGKDD Service Award, the 2014 IEEE ICDM Research Contributions Award, the British Columbia Innovation Council 2005 Young Innovator Award, an NSERC 2008 Discovery Accelerator Supplements Award, an IBM Faculty Award (2006), a KDD Best Application Paper Award (2008), an ICDE Influential Paper Award (2018), a PAKDD Best Paper Award (2014), and a PAKDD Most Influential Paper Award (2009).
\end{IEEEbiography}

\begin{IEEEbiography}[{\includegraphics[width=1in,height=1.3in,clip,keepaspectratio]{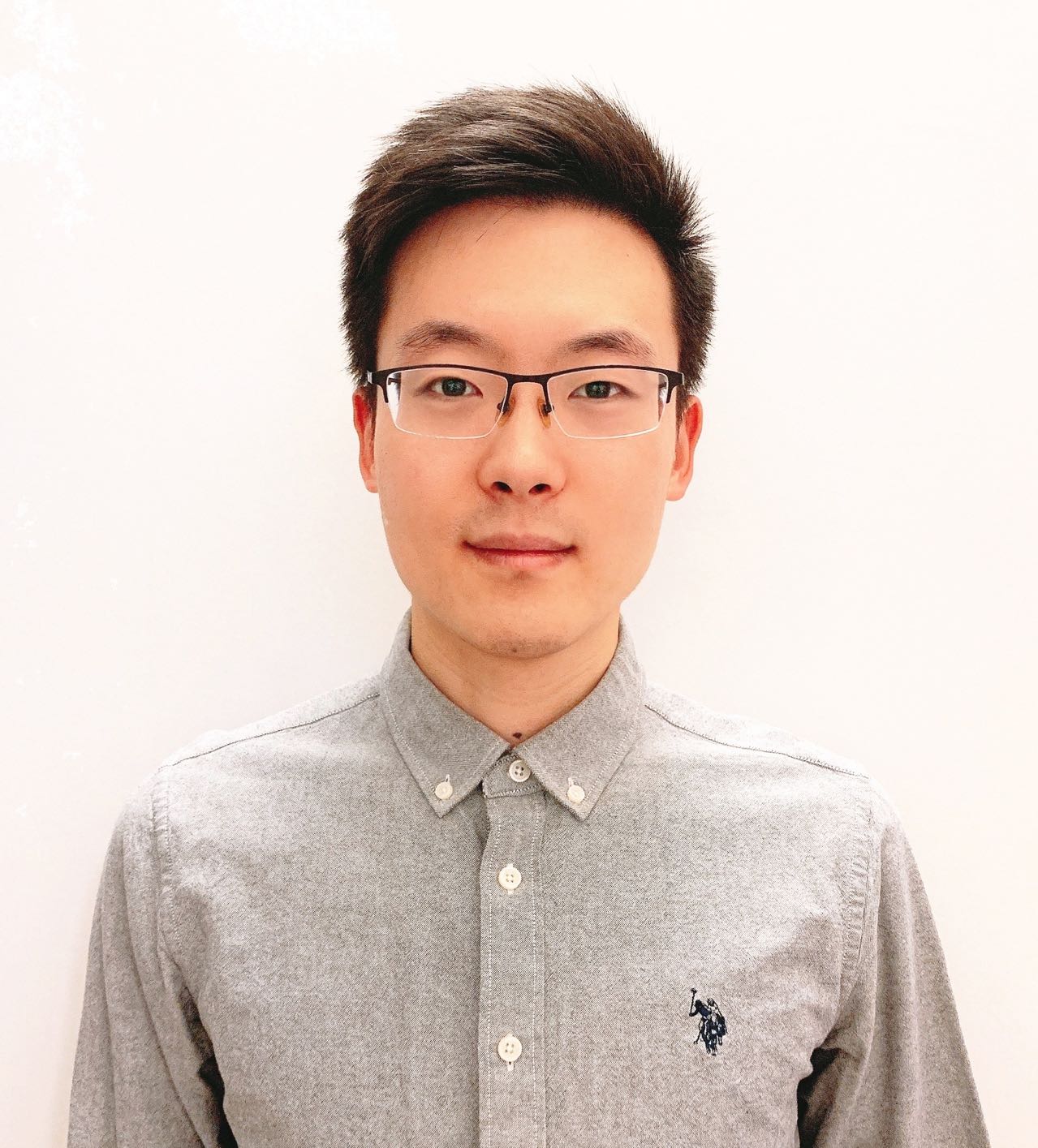}}]{Bo Zhang} is a researcher of the Search Product Center, WeChat Recommend Product Department, Tencent Inc., China. He got his BEng degree in 2009 from School of Computer Science, Xidian University, China, and got his Master Degree in 2012 from School of Computer Science and Technology, Zhejiang University, China.
\end{IEEEbiography}

\begin{IEEEbiography}[{\includegraphics[width=1in,height=1.3in,clip,keepaspectratio]{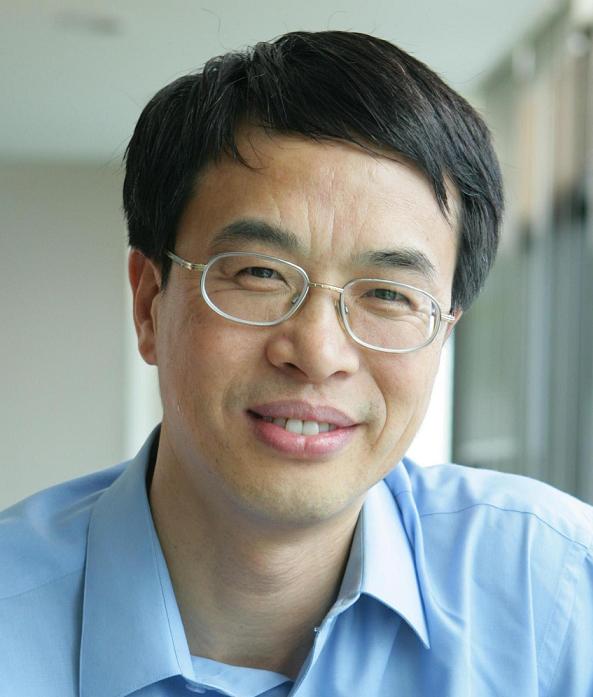}}]{Wenwu Zhu} is currently a Professor of the Computer Science Department of Tsinghua University. Prior to his current post, he was a Senior Researcher and Research Manager at Microsoft Research Asia. He was the Chief Scientist and Director at Intel Research China from 2004 to 2008. He worked at Bell Labs New Jersey as a Member of Technical Staff during 1996-1999. He received his Ph.D. degree from New York University in 1996.
	
He served as the Editor-in-Chief for the IEEE Transactions on Multimedia (T-MM) from January 1, 2017, to December 31, 2019. He has been serving as Vice EiC for IEEE Transactions on Circuits and Systems for Video Technology (TCSVT) and the chair of the steering committee for IEEE T-MM since January 1, 2020. His current research interests are in the areas of multimedia computing and networking, and big data. He has published over 350 papers in the referred journals and received nine Best Paper Awards including IEEE TCSVT in 2001 and 2019, and ACM Multimedia 2012. He is an IEEE Fellow, AAAS Fellow, SPIE Fellow and a member of the European Academy of Sciences (Academia Europaea).
\end{IEEEbiography}


\appendices
\clearpage
\section{Additional Experimental Results}\label{sec:exp_addresults}

\subsection{Comparison with Using Node IDs}\label{sec:exp_addresults_ID}
We compare SMP with augmenting GNNs using a one-hot encoding of node IDs, i.e., the identity matrix. Intuitively, since the IDs of nodes are unique, such a method does not suffer from the automorphism problem and should also enable GNNs to preserve node proximities. However, using such a one-hot encoding has two major problems. First, the dimensionality of the identity matrix is $N \times N$. Thus, the number of parameters in the first message-passing layer is also $O(N)$. Therefore, the method is inevitably computationally expensive and may not be scalable to large-scale graphs. Having a large number of parameters may also cause overfitting. Second, the node IDs are not transferable across different graphs, i.e., the node with ID $v_1$ in one graph and the node with ID $v_1$ in another graph do not necessarily share any similarity. Since the parameters in the message-passing depend on the node IDs as input features, using one-hot encoding cannot handle inductive tasks well.

We empirically compare such a method with SMP and report the results in Table~\ref{tab:exp-onehot}. The results show that SMP-Linear outperforms GCN$_\text{onehot}$ in most cases. Besides, GCN$_\text{onehot}$ fails to handle Physics, which is only a medium-scale graph, due to the heavy memory usage. One surprising result is that GCN$_\text{onehot}$ outperforms SMP-Linear on Grid, the simulated graph where nodes are placed on a $20 \times 20$ grid. A plausible reason is that since the edges in Grid follow a specific rule, using a one-hot encoding gives GCN$_\text{onehot}$ enough flexibility to learn the rules, and the model does not overfit because the graph has a rather small scale.

One may wonder whether SMP is transferable across different graphs, since the stochastic features are independently drawn. Empirically, we find that SMP reports reasonably good performance on inductive datasets, such as Email and PPI. One plausible reason is that, since the proximities of nodes are preserved even the random features per se are different (see Theorem~\ref{thm:walksim2} and Theorem~\ref{thm:walksim3}) , all subsequent parameters based on proximities can be transferred.

Besides, only using the stochastic representation of SMP in Eq.~\eqref{eq:stochasticrep} can be regarded as combining node IDs while fixing the first-layer parameters during training (assuming the first message-passing layer takes the form of matrix multiplication between node features and the weight matrix, as in SGC, GCN, and GAT). By fixing the parameters, both the computational bottlenecks and transferability problems are resolved.  

\begin{table*}[t]
\centering
\caption{The results of comparing SMP with using one-hot node IDs. OOM indicates out of memory. ``---'' indicates that we do not adopt the dataset for the task because the dataset does not have ground-truth labels.}
\begin{tabular}{@{}c|c c c c c c c c c c@{}}
\toprule
Task & Model & Grid         & Comm  & Email        & CS           & Physics      & PPI          & Cora         & CiteSeer     & PubMed       \\ \midrule
Link & GCN$_\text{onehot}$  & 91.5$\pm$2.1 & 98.3$\pm$0.7 & 71.2$\pm$3.5 & 93.1$\pm$1.3 & OOM          & 78.6$\pm$0.3 & 86.8$\pm$1.5 & 81.7$\pm$1.1 & 89.4$\pm$0.5 \\
Prediction     & SMP-Linear & 73.6$\pm$6.2 & 97.7$\pm$0.5 & 75.7$\pm$5.0 & 96.7$\pm$0.1 & 96.1$\pm$0.1 & 81.9$\pm$0.3 & 92.7$\pm$0.7 & 92.6$\pm$1.0 & 95.4$\pm$0.2 \\ \midrule
Pairwise Node  & GCN$_\text{onehot}$ &---  & 98.9$\pm$0.5 & 67.3$\pm$5.6 & 97.6$\pm$0.2 & OOM          & ---          & 98.2$\pm$0.3 & 94.4$\pm$1.2 & 98.9$\pm$0.1 \\
Classification & SMP-Linear & ---          & 98.8$\pm$0.5 & 74.5$\pm$4.1 & 99.8$\pm$0.0 & 99.6$\pm$0.0 & ---          & 99.3$\pm$0.3 & 95.3$\pm$0.4 & 93.4$\pm$0.2 \\ \midrule
Node           & GCN$_\text{onehot}$ &---  & 99.6$\pm$1.0 & ---          & 86.9$\pm$1.5 & OOM          & ---          & 77.6$\pm$1.1 & 57.7$\pm$5.8 & 74.9$\pm$0.6 \\
Classification & SMP-Linear & ---          & 99.9$\pm$0.3 & ---          & 91.5$\pm$0.8 & 93.1$\pm$0.8 & ---          & 80.9$\pm$0.8 & 68.2$\pm$1.0 & 76.5$\pm$0.8 \\ \bottomrule
\end{tabular}
\label{tab:exp-onehot}
\end{table*}

\subsection{Additional Link Prediction Results}\label{sec:exp_addresults_LP}
We further report the link prediction results on three GNN benchmarks: Cora, CiteSeer, and PubMed. The results in Table~\ref{tab:exp-link-add} show similar trends as other datasets presented in Section~\ref{sec:exp-linkprediction}, i.e., SMP reports comparable results to the other permutation-equivariant GNNs while P-GNN cannot handle the task well.

\begin{table}[ht]
\centering
\caption{The results of link prediction measured in AUC (\%) on three benchmarks. The best result and the second-best result for each dataset, respectively, is in bold and underlined.}
\begin{tabular}{@{}l|lll@{}}
\toprule
Model                          & Cora                  & CiteSeer              & PubMed                \\ \midrule
SGC                            & \textbf{93.6$\pm$0.6} & \textbf{94.7$\pm$0.8} & \textbf{95.8$\pm$0.2} \\
GCN                            & 90.6$\pm$1.0          & 78.2$\pm$1.7          & 92.4$\pm$0.9          \\
GAT                            & 88.5$\pm$1.2          & 87.8$\pm$1.0          & 89.2$\pm$0.8          \\ \midrule
P-GNN                           & 75.4$\pm$2.3          & 70.6$\pm$1.1          & Out of memory         \\ \midrule
SMP-Identity                   & {\ul 93.0$\pm$0.6}    & {\ul 92.9$\pm$0.5}    & 94.5$\pm$0.3          \\
SMP-Linear                     & 92.7$\pm$0.7          & 92.6$\pm$1.0          & {\ul 95.4$\pm$0.2}    \\
SMP-MLP                        & 82.8$\pm$0.9          & 80.7$\pm$1.1          & 88.0$\pm$0.6          \\
SMP-Linear-GCN$_{\text{feat}}$ & 86.7$\pm$1.4          & 81.1$\pm$1.4          & 90.5$\pm$0.6          \\
SMP-Linear-GCN$_{\text{both}}$ & 80.1$\pm$2.5          & 80.0$\pm$2.0          & 81.1$\pm$2.0
\\ \bottomrule
\end{tabular}
\label{tab:exp-link-add}
\end{table}

\subsection{Additional Ablation Studies}\label{sec:exp_addresults_ablation}
We report ablation study results for the node classification task and pairwise node classification task in Table~\ref{tab:exp-ablation2} and Table~\ref{tab:exp-ablation3}, respectively. The results again show that SMP-Linear generally achieves good-enough results on the majority of the datasets, and adding non-linearity does not consistently lift the performance.
\begin{table*}[ht]
\centering
\caption{The ablation study of different SMP variants for the node classification task. The best results and the second-best results are in bold and underlined, respectively.}
\begin{tabular}{@{}l|lll|lll@{}}
\toprule
Model          & Comm      & CS            & Physics                                           & Cora & CiteSeer & PubMed          \\ \midrule
SMP-Linear                     & 99.9$\pm$0.3           & \textbf{91.5$\pm$0.8} & \textbf{93.1$\pm$0.8} & \textbf{80.9$\pm$0.8} & \textbf{68.2$\pm$1.0} & 76.5$\pm$0.8          \\
SMP-MLP                        & \textbf{100.0$\pm$0.2} & {\ul 90.1$\pm$0.5}    & 92.3$\pm$0.8          & {\ul 79.3$\pm$0.8}    & 67.0$\pm$1.5          & {\ul 76.8$\pm$0.9}    \\
SMP-Linear-GCN$_{\text{feat}}$ & \textbf{100.0$\pm$0.0} & 89.8$\pm$0.7          & {\ul 92.9$\pm$0.8}    & 78.9$\pm$1.2          & {\ul 67.8$\pm$0.6}    & \textbf{77.3$\pm$0.6} \\
SMP-Linear-GCN$_{\text{both}}$ & \textbf{100.0$\pm$0.2} & 77.4$\pm$4.2          & 87.1$\pm$3.5          & 69.2$\pm$2.5          & 49.8$\pm$3.1          & 68.1$\pm$4.1
         \\ \bottomrule
\end{tabular}
\label{tab:exp-ablation2}
\end{table*}
\begin{table*}[ht]
\centering
\caption{The ablation study of different SMP variants for the pairwise node classification task. The best results and the second-best results are in bold and underlined, respectively.}
\begin{tabular}{@{}l|llll|lll@{}}
\toprule
Model                  & Comm             & Email                   & CS                   & Physics  & Cora & CiteSeer & PubMed              \\ \midrule
SMP-Identity                   & {\ul 98.8$\pm$0.5}    & 56.9$\pm$4.1          & {\ul 99.7$\pm$0.0}    & \textbf{99.6$\pm$0.0} & {\ul 99.2$\pm$0.2}    & {\ul 95.2$\pm$1.1} & 91.9$\pm$0.3          \\
SMP-Linear                     & {\ul 98.8$\pm$0.5}    & \textbf{74.5$\pm$4.1} & \textbf{99.8$\pm$0.0} & \textbf{99.6$\pm$0.0} & \textbf{99.3$\pm$0.3} & \textbf{95.3$\pm$0.4} & 93.4$\pm$0.2          \\
SMP-MLP                        & 98.7$\pm$0.3          & {\ul 65.4$\pm$6.3}    & 94.3$\pm$0.6          & 97.6$\pm$0.4          & 90.3$\pm$3.0          & 67.7$\pm$13.7         & 93.4$\pm$0.4          \\
SMP-Linear-GCN$_{\text{feat}}$ & \textbf{99.0$\pm$0.4} & 60.2$\pm$9.3          & 95.6$\pm$0.7          & \underline{98.3$\pm$0.7}          & 96.1$\pm$0.7          & 88.8$\pm$1.6          & \textbf{94.8$\pm$0.2} \\
SMP-Linear-GCN$_{\text{both}}$ & {\ul 98.8$\pm$0.4}    & 61.6$\pm$6.0          & 95.2$\pm$0.7          & 97.8$\pm$0.8          & 94.3$\pm$1.9          & 83.5$\pm$3.9          & {\ul 94.1$\pm$0.7}
\\\bottomrule
\end{tabular}
\label{tab:exp-ablation3}
\end{table*}

\subsection{Comparison with Position Encoding}\label{sec:exp_addresults_pos}

To compare the effectiveness of our proposed method with position encodings of GNNs, we adopt one recent method EigenGNN~\cite{zhang2021eigen}, which uses the eigenvectors of a graph structure matrix to enhance the existing GNNs in preserving graph structures. Specifically, we use GCN as the base architecture for EigenGNN. All hyper-parameters are kept consistently to ensure a fair comparison. The results for the link prediction task are shown in Table~\ref{tab:exp-pos}. Notice that we omit the results on PPA since calculating eigenvectors for extremely large-scale graphs is infeasible.

From the table, we can observe that SMP-Linear outperforms or is comparable to EigenGNN on eight of nine datasets, demonstrating the effectiveness of SMP. One exception is on Grid, where EigenGNN shows remarkably better results. We attribute this to the fact that Grid is an especially regular simulated graph (recall that nodes are placed on a $20 \times 20$ grid) and therefore EigenGNN can better capture this pattern. Besides, both EigenGNN and SMP greatly outperform GCN in most cases, demonstrating the importance of preserving proximities for GNNs.
\begin{table*}[t]
	\centering
	\caption{The results of comparing SMP with EigenGNN for the link prediction task measured by AUC. The best result for each dataset is in bold.}
	\begin{tabular}{@{}c|c c c c c c c c c c@{}}
		\toprule
		Model        & Grid         & Comm         & Email        & CS           & Physics      & PPI          & Cora         & CiteSeer     & PubMed       \\ \midrule
		GCN          & 61.8$\pm$3.6 & 50.3$\pm$2.5 & 67.4$\pm$6.9 & 93.4$\pm$0.3 & 93.8$\pm$0.2 & 78.0$\pm$0.4 & 90.6$\pm$1.0 & 78.2$\pm$1.7 & 92.4$\pm$0.9 \\
		EigenGNN     & \textbf{92.0$\pm$0.7} & \textbf{97.7$\pm$0.3} & 74.7$h\pm$3.1 & 93.2$\pm$0.3 & 93.4$\pm$0.8 & 81.7$\pm$0.4 & 88.7$\pm$0.9 & 83.0$\pm$1.3 & 91.0$\pm$0.3 \\ 
		SMP-Linear   & 73.6$\pm$6.2 & \textbf{97.7$\pm$0.5} & \textbf{75.7$\pm$5.0} & \textbf{96.7$\pm$0.1} & \textbf{96.1$\pm$0.1} & \textbf{81.9$\pm$0.3} & \textbf{92.7$\pm$0.7} & \textbf{92.6$\pm$1.0} & \textbf{95.4$\pm$0.2} \\
		\bottomrule
	\end{tabular}
	\label{tab:exp-pos}
\end{table*}

\section{Details for Reproducibility}
\subsection{Datasets}\label{sec:datasets}
\begin{itemize}[leftmargin=0.5cm]
	\item Proof-of-concept synthetic dataset: the dataset is generated using an add-on version of the stochastic block model~\cite{newman2018networks}. The graph has 400 nodes in 10 communities. Each node has an independent equal chance to be active or inactive. The probability of forming edges is (1) 0.8; (2) 0.5; (3) 0.2; and (4) 0.005, respectively, between (1) two active nodes in the same community; (2) an active node and an inactive one in the same community; (3) two inactive nodes in the same community; and (4) two nodes in different communities. There is no node feature.
	\item \textbf{Grid}~\cite{you2019position}: A simulated 2D grid graph with size $20 \times 20$.
	\item \textbf{Communities}~\cite{you2019position}: A simulated caveman graph~\cite{watts1999networks} composed of 20 communities, each community containing 20 nodes. The graph is perturbed by rewiring 1\% edges randomly. It has no node feature. The label of each node indicates the community that the node belongs to.
	\item \textbf{Email}\footnote{\label{url:pgnndata}\url{https://github.com/JiaxuanYou/P-GNN/tree/master/data}}~\cite{you2019position}: Seven real-world email communication graphs. Each graph has six communities and each node has an integer label indicating the community that the node belongs to.
	\item \textbf{Coauthor Networks}\footnote{\url{https://github.com/shchur/gnn-benchmark/tree/master/data/npz/}}~\cite{shchur2018pitfalls}: Two networks from Microsoft academic graph in CS and Physics with nodes representing authors and edges representing co-authorships. The node features are embeddings of the paper keywords of the authors.
	\item \textbf{PPI}\textsuperscript{\ref{url:pgnndata}}~\cite{hamilton2017inductive}: 24 protein-protein interaction networks. Each node has a 50-dimensional feature vector. 
	\item \textbf{PPA}\footnote{\url{https://snap.stanford.edu/ogb/data/linkproppred/ppassoc.zip}}~\cite{hu2020open}: A network representing biological associations between proteins from 58 species. The node features are one-hot vectors of the species that the proteins are taken from. 
	\item \textbf{Cora}, \textbf{CiteSeer}, \textbf{PubMed}\footnote{\url{https://github.com/kimiyoung/planetoid/tree/master/data}}~\cite{sen2008collective}: Three citation graphs where the nodes correspond to papers and the edges correspond to citations between papers. The node features are bag-of-words and the node labels are the ground truth topics of the papers.
\end{itemize}

\subsection{Hyper-parameters}\label{sec:hyperp}
\begin{itemize}[leftmargin=0.5cm]
	\item All datasets except PPA: we uniformly set the number of hidden layers to $2$ for all methods, i.e., two message-passing steps, and set the dimensionality of hidden layers to 32, i.e., $\mathbf{H}^{(l)} \in \mathbb{R}^{N\times 32}$, for $1 \leq l \leq L$ (for GAT, we use 4 heads with each head containing 8 units).
	We use the Adam optimizer with an initial learning rate of 0.01 and decay the learning rate by 0.1 at epoch 200. The weight decay is 5e-4.  We train the model for 1,000 epochs and evaluate the model every 5 epochs. We report the testing performance at the epoch which achieves the best validation performance. For SMP, the dimensionality of the stochastic matrix is $d=32$. For P-GNN, we use the P-GNN-F version, which uses the truncated 2-hop shortest path distance instead of the exact shortest distance for scalability.
	\item PPA: as suggested in the original paper~\cite{hu2020open}, we set the number of GNN layers to 3 with each layer containing 256 hidden units, and add a three-layer MLP after taking the Hadamard product between pair-wise node embeddings as the predictor, i.e., $\operatorname{MLP}(\mathbf{H}_i \odot \mathbf{H}_j)$. We use the Adam optimizer with an initial learning rate of 0.01. We set the number of epochs for training to 40, evaluate the results on validation sets every epoch, and report the testing results using the model with the best validation performance. We find that the dataset has issues with exploding gradients, and thus adopt a gradient clipping strategy by limiting the maximum
	p2-norm of gradients to $1.0$.
	The dimensionality of the stochastic matrix in SMP is $d=64$.
\end{itemize}

\subsection{Hardware and Software Configurations}
All experiments are conducted on a server with the following configurations.
\begin{itemize}
\item Operating System: Ubuntu 18.04.1 LTS
\item CPU: Intel(R) Xeon(R) CPU E5-2699 v4 @ 2.20GHz
\item GPU: NVIDIA TESLA M40 with 12 GB of memory
\item Software: Python 3.6.8, PyTorch 1.4.0, PyTorch Geometric 1.4.3, NumPy 1.18.1, Cuda 10.1
\end{itemize}

\section{Proof}
\subsection{Proof of Theorem~\ref{thm:walksim3}}\label{sec:sophi}
\begin{proof}
	Similar to the proof of Theorem~\ref{thm:walksim2}, we only utilize $\tilde{\mathbf{E}}$ in our proof. Denote by $\mathbf{e}^{(l)}_i, 0 \leq l \leq L$, the representations of node $v_i$ in the $l^{th}$ layer of  $\mathcal{F}_{\text{GNN}}\left(\mathbf{A},\mathbf{E}; \mathbf{W} \right)$, i.e., $\mathbf{e}^{(0)}_i = \mathbf{E}_{i,:}$ and $\mathbf{e}^{(L)}_i = \tilde{\mathbf{E}}_{i,:}$. Our proof strategy is to show that the stochastic node representations can remember all the information about the walks, which can be decoded by the decoder $\mathcal{F}_{\text{de}}(\cdot)$.
	
	First, as the message-passing and the updating function are bijective by assumption, we can recover from the node representations in each layer all their neighborhood representations in the previous layer. Specifically, there exists $\mathcal{F}^{(l)}(\cdot)$, $1\leq l \leq L$, such that
	\begin{equation}
		\mathcal{F}^{(l)}\left(\mathbf{e}^{(l)}_i\right) = \left[ \mathbf{e}^{(l-1)}_i, \left\{ \mathbf{e}^{(l-1)}_j, j \in \mathcal{N}_i \right\} \right]
		\footnote{To let $\mathcal{F}^{(l)}(\cdot)$ output a set with arbitrary lengths, we can adopt sequence-based models such an LSTM.}.
	\end{equation}
	
	To keep our notations clear, we split the function into two parts, one for the node itself and the other one for its neighbors
	\begin{equation}
		\begin{gathered}
			\mathcal{F}_{\text{self}}^{(l)}\left(\mathbf{e}^{(l)}_i\right) = \mathbf{e}^{(l-1)}_i,
			\mathcal{F}_{\text{neighbor}}^{(l)}\left(\mathbf{e}^{(l)}_i\right) = \left\{ \mathbf{e}^{(l-1)}_j, j \in \mathcal{N}_i \right\}.
		\end{gathered}
	\end{equation}
	For the first function, if we successively apply such functions from the $l^{th}$ layer to the input layer, we can recover the input features of the GNN, i.e., $\mathbf{E}$. Since the stochastic matrix $\mathbf{E}$ contains an identifiable unique signal for each node, we can decode the node ID from $\mathbf{e}^{(0)}_i$, i.e., there exists $\mathcal{F}_{\text{self}}^{(0)}\left(\mathbf{e}^{(0)}_i;\mathbf{E}\right) = i$. We denote the process of applying such $l+1$ functions to get the node ID as
	\begin{equation}
		\mathcal{F}_{\text{self}}^{(0:l)}\left( \mathbf{e}_i^{(l)}\right) = \mathcal{F}_{\text{self}}^{(0)}\left( \mathcal{F}_{\text{self}}^{(1)}\left( ... \left( \mathcal{F}_{\text{self}}^{(l)}\left( \mathbf{e}_i^{(l)}\right) \right) \right);\mathbf{E}\right) = i.
	\end{equation}
	For the second function, we can apply $\mathcal{F}_{\text{neighbor}}^{(l-1)}(\cdot)$ to the decoded vector set so that we can recover their neighborhood representations in the $(l-2)^{th}$ layer. Similar procedures can be performed successively until we reach the first layer.
	
	Next, we show that for $\mathbf{e}^{(l)}_{j}$, there exists a length-$l$ walk $v_i \leadsto v_j = (v_{a_1},v_{a_2},...,v_{a_{l+1}})$, where $v_{a_1} = v_i$, $v_{a_{l+1}}=v_j$, if and only if $\mathcal{F}_{\text{self}}^{(0:l)}\left( \mathbf{e}_j^{(l)}\right) = a_l = j$ and there exist $\mathbf{e}^{(l-1)},...,\mathbf{e}^{(0)}$ such that:
	\begin{equation}
		\begin{gathered}
			\mathbf{e}^{(l-1)} \in \mathcal{F}_{\text{neighbor}}^{(l)}\left( \mathbf{e}^{(l)}_{j} \right),  \mathcal{F}_{\text{self}}^{(0:l-1)}\left( \mathbf{e}^{(l-1)}\right) = a_{l},\\
			\mathbf{e}^{(l-2)} \in \mathcal{F}_{\text{neighbor}}^{(l-1)}\left( \mathbf{e}^{(l-1)} \right),  \mathcal{F}_{\text{self}}^{(0:l-2)}\left( \mathbf{e}^{(l-2)}\right) = a_{l-1}, \\
			... \\
			\mathbf{e}^{(0)} \in \mathcal{F}_{\text{neighbor}}^{(1)}\left( \mathbf{e}^{(1)} \right),  \mathcal{F}_{\text{self}}^{(0:0)}\left( \mathbf{e}^{(0)}\right) = a_1 = i.
		\end{gathered}
	\end{equation}
	
	This result is easy to obtain as follows. 
	\begin{small}
	\begin{displaymath}
		\begin{aligned}
			& (v_{a_1},v_{a_2},...,v_{a_{l+1}}) \text{ is a walk}  \\
			\Leftrightarrow & \mathcal{E}_{a_i,a_{i+1}} = \mathcal{E}_{a_{i+1}, a_i} = 1, \forall 1 \leq i \leq l
			\Leftrightarrow a_{i} \in \mathcal{N}_{a_{i+1}}, \forall 1 \leq i \leq l \\
			\Leftrightarrow & \exists \mathbf{e}^{(i-1)} \in \mathcal{F}_{\text{neighbor}}^{(i)}\left( \mathbf{e}^{(i)} \right),  \mathcal{F}_{\text{self}}^{(0:i-1)}\left( \mathbf{e}^{(i-1)}\right) = a_{i}, \forall 1 \leq i \leq l .
		\end{aligned}
	\end{displaymath}
	\end{small}
	
	Note that all the information is encoded in $\tilde{\mathbf{E}}$, i.e., we can decode $\left\{v_i \leadsto v_j | \text{len}(v_i \leadsto v_j) \leq L\right\}$ from $\mathbf{e}_j^{(L)}$ by successively applying $\mathcal{F}_{\text{self}}^{(l)}\left(\cdot\right)$ and $\mathcal{F}_{\text{neighbor}}^{(l)}\left(\cdot\right)$. We can also apply $\mathcal{F}_{\text{self}}^{(0:L)}$ to $\mathbf{e}_i^{(L)}$ to get the start node ID $i$. Putting all together, we have
	\begin{equation}
		\mathcal{F}\left(\mathbf{e}_j^{(L)},\mathbf{e}_i^{(L)}\right) = \left\{v_i \leadsto v_j | \text{len}(v_i \leadsto v_j) \leq L\right\},
	\end{equation}
	where $\mathcal{F}(\cdot)$ is composed of $\mathcal{F}_{\text{self}}^{(l)}\left(\cdot\right)$, $0\leq l \leq L$, and $\mathcal{F}_{\text{neighbor}}^{(l)}\left(\cdot\right)$, $1\leq l \leq L$. Applying the proximity function $\mathcal{S}(\cdot)$, we have:
	\begin{equation}
		\mathcal{S} \left( \mathcal{F}\left(\mathbf{e}_j^{(L)},\mathbf{e}_i^{(L)}\right) \right) = \mathbf{S}_{i,j}.
	\end{equation}
	We finish the proof by setting the real decoder function $\mathcal{F}_{\text{de}}(\cdot)$ to arbitrarily approximate this desired function $\mathcal{S} \left( \mathcal{F}\left(\cdot,\cdot\right) \right)$ under the universal approximation assumption.
\end{proof}

\subsection{An Alternative Proof of Theorem~\ref{thm:walksim}}\label{sec:proof:alter}
Some may find that our proof of Theorem~\ref{thm:walksim} in the main paper leads to multiple connected components in the constructed graph $G^\prime$. Next, we give an alternative proof maintaining one connected component in $G^\prime$ assuming the original graph $G$ is connected) under an additional assumption that the walk-based proximity is of finite length.

\begin{proof}
	Similar to the previous proof, we will prove by contraction. We assume there exists a non-trivial $\mathcal{S}(\cdot)$ that a permutation-equivariant GNN can preserve. Besides, we assume the length of $\mathcal{S}(\cdot)$ is upper bounded by $l_{\text{max}}$, where $l_{\text{max}}$ is any finite number, i.e., $\forall i,j$,
	\begin{equation}
		\mathbf{S}_{i,j} = \mathcal{S}\left (\left\{v_i \leadsto v_j \right\} \right) = \mathcal{S}\left (\left\{v_i \leadsto v_j | \text{len}(v_i \leadsto v_j) \leq l_{\text{max}} \right\} \right),
	\end{equation}
	where $\text{len}(v_i \leadsto v_j)$ is the length, i.e., number of nodes minus 1, of the walk $v_i \leadsto v_j$.
	
	For a connected graph $G=\left(\mathcal{V},\mathcal{E},\mathbf{F}\right)$, we create $G^\prime=\left(\mathcal{V}^\prime,\mathcal{E}^\prime,\mathbf{F}^\prime\right)$ that has automorphism. Then we will show that any GNN cannot preserve $\mathcal{S}(\cdot)$ on $G^\prime$. Specifically, denoting $\tilde{N} = N + l_{\text{max}}$, we let $G^\prime$ have $3\tilde{N}$ nodes so that:
	\begin{small}
		\begin{equation}\label{eq:construct2}
			\begin{gathered}
				\mathcal{E}^\prime_{i,j} = \left\{
				\begin{aligned}
					& \mathcal{E}_{i,j} &\text{if } i, j\leq N \\
					& 1                 &\text{if } N \leq i,j \leq \tilde{N} +1, \left|j - i\right| = 1 \\
					& \mathcal{E}_{i-\tilde{N},j-\tilde{N}} &\text{if } \tilde{N} < i,j \leq \tilde{N} + N\\
					& 1                 &\text{if } \tilde{N} + N \leq i,j \leq 2\tilde{N} +1, \left|j - i\right| = 1 \\
					& \mathcal{E}_{i-2\tilde{N},j-2\tilde{N}} &\text{if } 2\tilde{N} < i,j \leq 2\tilde{N} + N \\
					& 1                 &\text{if } 2\tilde{N} +N \leq i,j, \left|j - i\right| = 1 \\
					& 1                 &\text{if } i=3\tilde{N},j=1 \text{ or } j=3\tilde{N},i=1 \\
					& 0                 &\text{else } \\
				\end{aligned} \right.,\\
				\mathbf{F}^\prime_{i,:} = \left\{
				\begin{aligned}
					& \mathbf{F}_{i,:} &\text{ if } i\leq N \\
					& 0  &\text{ if } N < i \leq \tilde{N} \\
					& \mathbf{F}_{i - \tilde{N},:} &\text{ if } \tilde{N} < i \leq \tilde{N} + N \\
					& 0  &\text{ if } \tilde{N} + N < i \leq 2 \tilde{N} \\
					& \mathbf{F}_{i - 2\tilde{N},:} &\text{ if } 2\tilde{N} < i \leq 2\tilde{N} + N \\
					& 0  &\text{ if } 2\tilde{N} + N < i
				\end{aligned} \right..
			\end{gathered}
		\end{equation}
	\end{small}Intuitively, we create three ``copies'' of $G$ and three ``bridges'' to connect the copies and thus make $G^\prime$ also connected. It is also easy to see that nodes $v_i^\prime$, $v_{i+\tilde{N}}^\prime$, and $v_{i+2\tilde{N}}^\prime$ all form automorphic node pairs. Therefore, we have:
	\begin{equation}
		\mathbf{H}^{\prime(L)}_{i,:} = \mathbf{H}^{\prime(L)}_{i + \tilde{N},:} = \mathbf{H}^{\prime(L)}_{i + \tilde{2N},:}, \forall i \leq \tilde{N}.
	\end{equation}
	
	Next, we see that the nodes in $G^\prime$ are divided into six parts (three copies and three bridges), which we denote as $\mathcal{V}^\prime_1 = \left\{v_1...v_N\right\}$,  $\mathcal{V}^\prime_2 = \left\{v_{N+1}...v_{\tilde{N}}\right\}$, $\mathcal{V}^\prime_3 = \left\{v_{\tilde{N}+1}...v_{\tilde{N}+N}\right\}$, $\mathcal{V}^\prime_4 = \left\{v_{\tilde{N}+N+1},...,v_{2\tilde{N}}\right\}$, $\mathcal{V}^\prime_5 = \left\{v_{2\tilde{N}+1},...,v_{2\tilde{N}+N}\right\}$, and $\mathcal{V}^\prime_6 = \left\{v_{2\tilde{N}+N+1},...,v_{3\tilde{N}}\right\}$. Since $\mathcal{V}^\prime_2$, $\mathcal{V}^\prime_4$, $\mathcal{V}^\prime_6$ are bridges with a length $l_{\text{max}}$, any walk crosses these bridges will have a length large than $l_{\text{max}}$. For example, let us focus on $v_i \in \mathcal{V}^\prime_1$, i.e., $1 \leq  i \leq N$. If $v_j$ is in $\mathcal{V}^\prime_3$, $\mathcal{V}^\prime_4$, or $\mathcal{V}^\prime_5$ (i.e., $\tilde{N} < j \leq 2\tilde{N} + N $), any walk $v_i \leadsto v_j$ will either cross the bridge $\mathcal{V}^\prime_2$ or $\mathcal{V}^\prime_6$ and thus has a length larger than $l_{\text{max}}$. As a result, we have:
	\begin{footnotesize}
		\begin{equation}
			\begin{gathered}
				\mathbf{S}_{i,j} = \mathcal{S}\left (\left\{v_i \leadsto v_j \right\} \right) = \mathcal{S}\left (\left\{v_i \leadsto v_j | \text{len}(v_i \leadsto v_j) \leq l_{\text{max}} \right\} \right) = \mathcal{S}\left( \emptyset\right).
			\end{gathered}
		\end{equation}
	\end{footnotesize}
	If $v_j \in \mathcal{V}^\prime_1$ or $v_j \in \mathcal{V}^\prime_2$, i.e., $j \leq \tilde{N}$, we can use the fact that $v_j$ and $v_{j+\tilde{N}}$ forms an automorphic node pair, i.e., $\forall \epsilon > 0$, we have
	\begin{scriptsize}
		\begin{equation}\label{eq:disinequality2}
			\begin{split}
				\left| \mathbf{S}_{i,j} - \mathcal{S}(\emptyset) \right|  \leq \left| \mathbf{S}_{i,j} -  \mathcal{F}_{\text{de}}\left(\mathbf{H}^{\prime(L)}_{i,:},\mathbf{H}^{\prime(L)}_{j,:}\right) \right| + \left| \mathcal{S}(\emptyset) - \mathcal{F}_{\text{de}}\left(\mathbf{H}^{\prime(L)}_{i,:},\mathbf{H}^{\prime(L)}_{j,:}\right) \right| \\
				= \left| \mathbf{S}_{i,j} -  \mathcal{F}_{\text{de}}\left(\mathbf{H}^{\prime(L)}_{i,:},\mathbf{H}^{\prime(L)}_{j,:}\right) \right| + \left| \mathbf{S}_{i,j+\tilde{N}} - \mathcal{F}_{\text{de}}\left(\mathbf{H}^{\prime(L)}_{i,:},\mathbf{H}^{\prime(L)}_{j+\tilde{N},:}\right) \right| < 2\epsilon.
			\end{split}
		\end{equation}
	\end{scriptsize}
	Similarly, if $v_j \in \mathcal{V}^\prime_6$, i.e., $2\tilde{N} + N < j$, we can use the fact that $v_j$ and $v_{j-\tilde{N}}$ forms an automorphic node pair to prove the same inequality. Thus, we prove that if $i \leq N, \forall \epsilon > 0, \forall j$,  $\left| \mathbf{S}_{i,j} - \mathcal{S}(\emptyset) \right| < 2 \epsilon$. The same proof strategy can be applied to $i > N$. Since $\epsilon$ can be arbitrarily small, the results show that all node pairs have the same proximity $\mathcal{S}(\emptyset)$, which leads to a contraction and finishes our proof.
\end{proof}

\end{sloppy}

\end{document}